\documentclass[10pt,twocolumn,letterpaper]{article}

\usepackage[pagenumbers]{template}           
\definecolor{cvprblue}{rgb}{0.21,0.49,0.74}
\usepackage[pagebackref,breaklinks,colorlinks,allcolors=cvprblue]{hyperref}
\usepackage{array}
\usepackage{multirow}
\usepackage{arydshln}
\usepackage{booktabs}       
\usepackage{amsfonts}       
\usepackage{nicefrac}       
\usepackage{microtype}      
\usepackage{xcolor}         
\usepackage{graphicx}
\usepackage{float}
\usepackage{amsmath}
\usepackage{makecell}
\usepackage[normalem]{ulem}
\usepackage{lipsum}
\usepackage{wrapfig}
\usepackage{tabularray}
\usepackage{verbatim}

\title{Divot: Diffusion Powers Video Tokenizer for Comprehension and Generation}

\author{
\textbf{Yuying Ge \qquad Yizhuo Li \qquad Yixiao Ge \qquad Ying Shan}\\
ARC Lab, Tencent PCG\\
\url{https://github.com/TencentARC/Divot}
}

\begin{document}
\maketitle
\begin{abstract}
In recent years, there has been a significant surge of interest in unifying image comprehension and generation within Large Language Models (LLMs). This growing interest has prompted us to explore extending this unification to videos. The core challenge lies in developing a versatile video tokenizer that captures both the spatial characteristics and temporal dynamics of videos to obtain representations for LLMs, and the representations can be further decoded into realistic video clips to enable video generation. 
In this work, we introduce Divot, a \textbf{Di}ffusion-Powered \textbf{V}ide\textbf{o} \textbf{T}okenizer, which leverages the diffusion process for self-supervised video representation learning. We posit that if a video diffusion model can effectively de-noise video clips by taking the features of a video tokenizer as the condition, then the tokenizer has successfully captured robust spatial and temporal information. Additionally, the video diffusion model inherently functions as a de-tokenizer, decoding videos from their representations. 
Building upon the Divot tokenizer, we present Divot-LLM through video-to-text autoregression and text-to-video generation by modeling the distributions of continuous-valued Divot features with a Gaussian Mixture Model. Experimental results demonstrate that our diffusion-based video tokenizer, when integrated with a pre-trained LLM, achieves competitive performance across various video comprehension and generation benchmarks. The instruction tuned Divot-LLM also excels in video storytelling, generating interleaved narratives and corresponding videos.
Models and codes are available at \url{https://github.com/TencentARC/Divot}
\end{abstract}

\section{Introduction}
\label{sec:intro}
\begin{figure}
	\centering
	\includegraphics[width=1.0\linewidth]{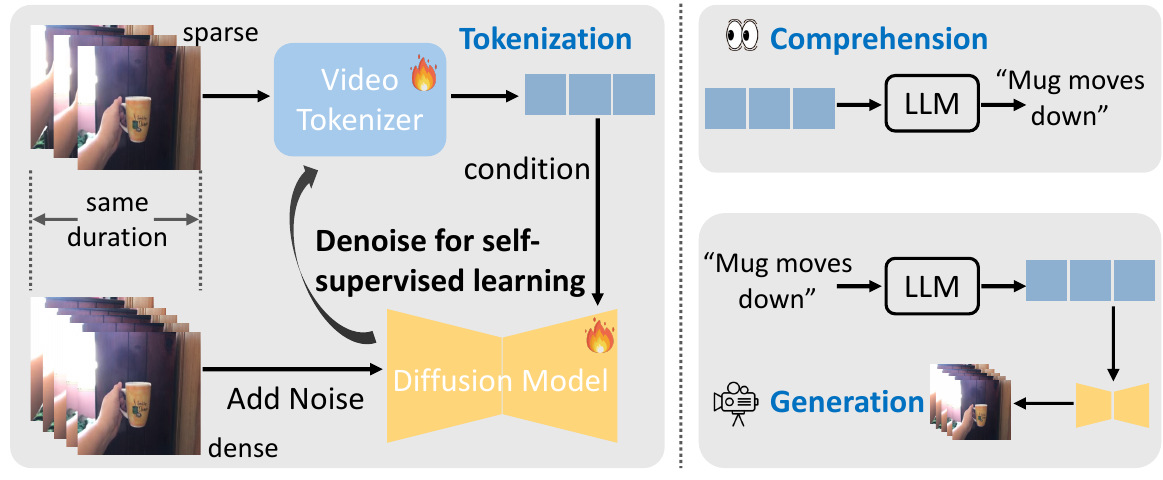}
	\caption{We utilize the \textbf{diffusion procedure} to learn a video tokenizer in a self-supervised manner for unified comprehension and generation, where the spatiotemporal representations serve as the condition of a diffusion model to de-noise video clips. Additionally, the proxy diffusion model functions as a de-tokenizer to decode realistic video clips from the video representations.}
	\label{fig:teaser}
	\vspace{-2mm}	
\end{figure}

In recent years, the rapid evolution of Multimodal Large Language Models (MLLMs)~\cite{sun2023emu, yu2023scaling, ge2023planting, ge2023making, dong2023dreamllm, sun2023generative, zhu2023vl, ge2024seed, xie2024show, team2024chameleon, zhou2024transfusion} has demonstrated significant
progresses in unified image understanding and generation, which empowers LLMs~\cite{touvron2023llama, brown2020language, chowdhery2022palm} with
the ability to generate images beyond texts. While these work primarily focus on image-text data, the extension of this unification to the video domain remains relatively under-explored. Achieving unified video comprehension and generation is essential for the development of more sophisticated artificial intelligence (AI) systems that are capable of understanding and creating dynamic visual content in the real world.

The primary challenge of achieving unified video comprehension and generation lies in developing a versatile video tokenizer that can effectively address the complexities inherent in video data. This tokenizer should be able to obtain robust video representations that serve as inputs of MLLMs for video comprehension, and these representations can be further decoded into realistic video clips to enable video generation. Unlike static images, videos encompass both spatial characteristics and temporal dynamics, making their representation significantly more complex. Recent pioneering work~\cite{liu2024world, jin2024video, wu2024vila} adopt a discrete video tokenizer for unifying video comprehension and generation, where a video is represented as a sequence of discrete frame tokens, or keyframe tokens followed by discrete motion tokens. This approach eases video generation with a LLM through an autoregressive next-token prediction mechanism, but sacrifices the performance of multimodal understanding, as pointed out by recent work~\cite{xie2024show}. In this work, we aim to investigate an alternative approach by \textit{utilizing continuous video representations} to unify video comprehension and generation.

To this end, we introduce Divot, a \textbf{Di}ffusion-Powered \textbf{V}ide\textbf{o} \textbf{T}okenizer that leverages the diffusion process~\cite{rombach2022high} for self-supervised video representation learning as shown in Fig.~\ref{fig:teaser}. The core premise is that if a diffusion model can effectively predict the noise added to the Variational Autoencoder (VAE) latents~\cite{kingma2013auto} of video clips, when conditioned on the features produced by the video tokenizer, it demonstrates that \textit{the tokenizer has successfully captured robust spatial and temporal information} inherent in the video data. This capability is crucial for representing the intricate dynamics present in videos. Furthermore, in addition to being a proxy module for learning the tokenizer, the diffusion model can act as a de-tokenizer to effectively decode realistic videos from their learned representations. This dual functionality facilitates a seamless integration of understanding and creating video content within a LLM.

Specifically, the Divot tokenizer is composed of a pre-trained Vision Transformer (ViT) encoder~\cite{dosovitskiy2020image}, a Spatial-Temporal transformer, and a Perceiver Resampler~\cite{alayrac2022flamingo} to obtain video representations from video frames sampled at low frame rate (fps) considering the semantic redundancy between adjacent frames. The video representations serve as the condition of a pre-trained video diffusion model, DynamiCrafter~\cite{xing2025dynamicrafter}  (without the concatenation of a conditional image with initial noise), to predict the noise added to the VAE latents of video frames. After training, the video diffusion model can generate realistic video clips from noise by taking the video representations provided by the Divot tokenizer as the condition.

We further present Divot-LLM by equipping the pre-trained Mistral-7B\footnote{We do not explore more advanced LLMs because we want to ensure that our superiority stems from the improved visual representations, rather than from the capabilities of a more sophisticated foundation model.}~\cite{jiang2023mistral} with the Divot tokenizer. Divot-LLM is pre-trained with a next-word prediction objective on video-caption data by taking the spatiotemporal representations of the Divot tokenizer as inputs for video comprehension. The challenge then arises in \textit{modeling the continuous video representations with the LLM} for video generation. We empirically find that simply minimizing the distance between the LLM output and video representations using mean squared error (MSE) loss achieves unsatisfactory results, since the deterministic regression regularizes the LLM to learn overly averaged features of videos. To address this, inspired by recent work~\cite{li2024autoregressive}, we shift our focus from deterministic modeling to \textbf{probabilistic modeling} by modeling the distributions of video features with a Gaussian Mixture Model (GMM). Specifically, we train the LLM to predict GMM parameters, including means, variances, and mixture probabilities through minimizing the discrepancy between the predicted GMM distribution and the actual video representations using negative log-likelihood (NLL) loss. During inference, we draw samples from the predicted GMM distribution as the condition of the video de-tokenizer to decode video clips.

We benchmark Divot-LLM on a broad range of video comprehension tasks and zero-shot video generation, achieving competitive performance through pre-training on 5 million video-text pairs using 32 A100-40G GPUs. By leveraging the generality of the video tokenizer, our Divot-LLM also enables video storytelling, which generates interleaved narratives and corresponding videos that are temporally coherent through fine-tuning on a specific animation dataset.

Our contributions are three-fold. (1) We introduce Divot, an advanced video tokenizer that leverages a diffusion procedure for self-supervised video representation learning, aiming to unify video comprehension and generation. (2) We present Divot-LLM, composed of a pre-trained LLM and the Divot tokenizer to enable understanding and generating video content within a single framework. We investigate effective approaches for fitting continuous video representations using the LLM with probabilistic modeling for video generation. (3) We conduct extensive experiments to demonstrate Divot-LLM's competitive performance on existing video comprehension and generation benchmarks, as well as video storytelling. All models and code are released.

\begin{figure*}
	\vspace{-4mm}	
	\centering
	\includegraphics[width=0.95\linewidth]{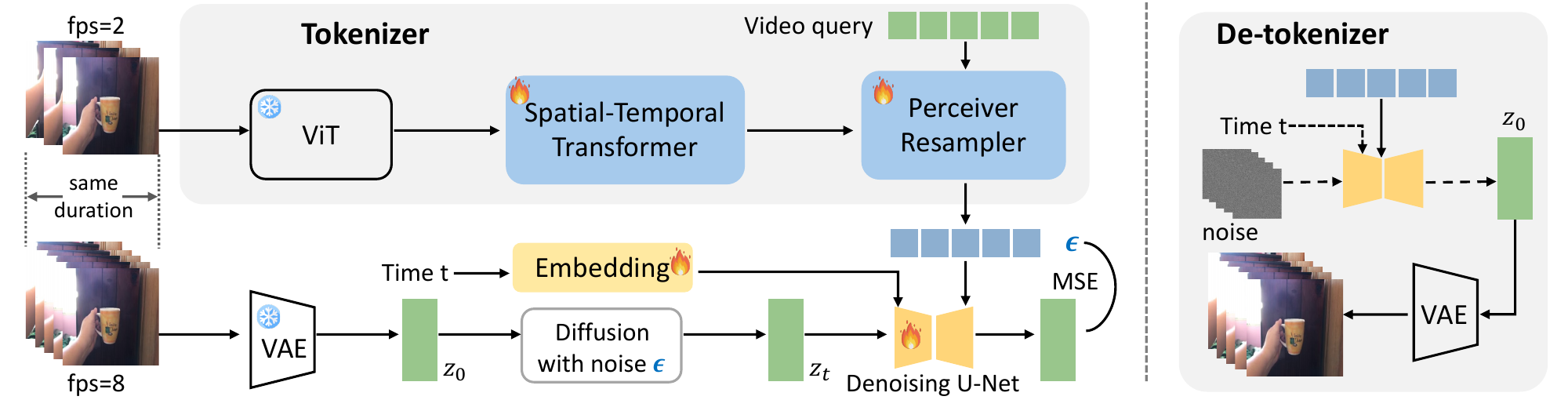}
	\caption{Overview of \textbf{Divot tokenization and de-tokenization}. During training, sparsely sampled video frames are fed into the tokenizer to obtain spatiotemporal representations. These representations serve as the conditions for a U-Net, which is trained to de-noise the noisy VAE latents of densely sampled video frames. During inference, the video representations from the Divot tokenizer can be decoded into realistic video clips with the U-Net.}
	\label{fig:method}
	\vspace{-2mm}	
\end{figure*}

\section{Related Work}
\paragraph{MLLMs for Comprehension and Generation.}
With the rapid development of Multimodal Large Language Models (MLLM), recent studies have been working on unified MLLMs~\citep{ge2023making, ge2023planting, yu2023scaling, sun2023emu, sun2023emu2, dong2023dreamllm, wu2023nextgpt, zhu2023vl, jin2023unified, lu2023unified, liu2024world, jin2024video, team2024chameleon, li2024mini, ge2024seed, yang2024seed, xie2024show, wu2024vila, zhou2024transfusion, zhao2024monoformer, wang2024emu3, wu2024janus, yang2024mmar} that are capable of multimodal comprehension and generation. To empower LLMs with the capability to generate visual content, existing work primarily employs the following three approaches: (1) utilizing a pre-trained stable diffusion model to generate images conditioned on LLM output (either continuous features or discrete tokens); (2) employing a Vector Quantized (VQ)~\cite{van2017neural} based decoder to generate visual content from the discrete codes predicted by LLMs; (3) using LLMs to de-noise Gaussian noise through a diffusion process. While most work predominantly focus on the unification of images and texts, some pioneering studies~\cite{liu2024world, jin2024video, wu2024vila, wang2024emu3} further advance the integration of video comprehension and generation within an LLM through generating videos from discrete codes using a VQ-based decoder, which falls into the second approach. In this work, we adopt the first approach, which involves leveraging a diffusion model to achieve unified video understanding and generation from continuous representations.

\paragraph{Video Tokenizer in MLLMs.}
Previous work on video generation with LLMs predominantly employs a discrete video tokenizer to convert video signals into a sequence of quantized tokens. For example, LWM~\cite{liu2024world} and VILA-U~\cite{wu2024vila} utilize a frame-level tokenizer to discretize each frame into a sequence of codes. VideoPoet~\cite{kondratyuk2023videopoet}, Loong~\cite{wang2024loong} and Emu3~\cite{wang2024emu3} leverage a 3D CNN architecture, where the encoded spatial-temporal features are quantized into discrete tokens. Video-LaVIT~\cite{jin2024video} represents video clips as a keyframe followed by extracted motion vectors, obtaining the respective discrete codes. By converting continuous visual signals into discrete tokens, the original next-token prediction mechanism can be adopted to facilitate video generation with an LLM. However, recent work~\cite{xie2024show} observes a significant performance degradation in multimodal comprehension tasks when discrete representations are used instead of continuous representations. In this work, we introduce a video tokenizer with continuous representations  through leveraging the diffusion~\cite{rombach2022high} procedure, enabling it to be effectively integrated with a LLM for unified comprehension and generation.

\paragraph{Diffusion for Representation Learning.}
The diffusion process has been explored as a criterion for representation learning. Some works~\cite{xiang2023denoising, baranchuk2021label, xu2023open, zhao2023unleashing} leverage the intermediate activations of pre-trained diffusion models for downstream tasks including classification, segmentation and depth estimation. Other works~\cite{wei2023diffusion, hudson2024soda, wang2024diffusion} employ the diffusion model as a proxy module for self-supervised learning, where noisy inputs are de-noised by conditioning on the image representations. This approach encourages the emergence of informative representations that capture key properties and semantics of the images. In this work, to the best of our knowledge, we \textbf{for the first time} leverage diffusion for video representation learning, where a video diffusion model is trained to de-noise video clips through taking the spatiotemporal representations as conditions, thereby encouraging the capture of spatial characteristics and temporal dynamics.

\begin{figure}
	\centering
	\includegraphics[width=1.0\linewidth]{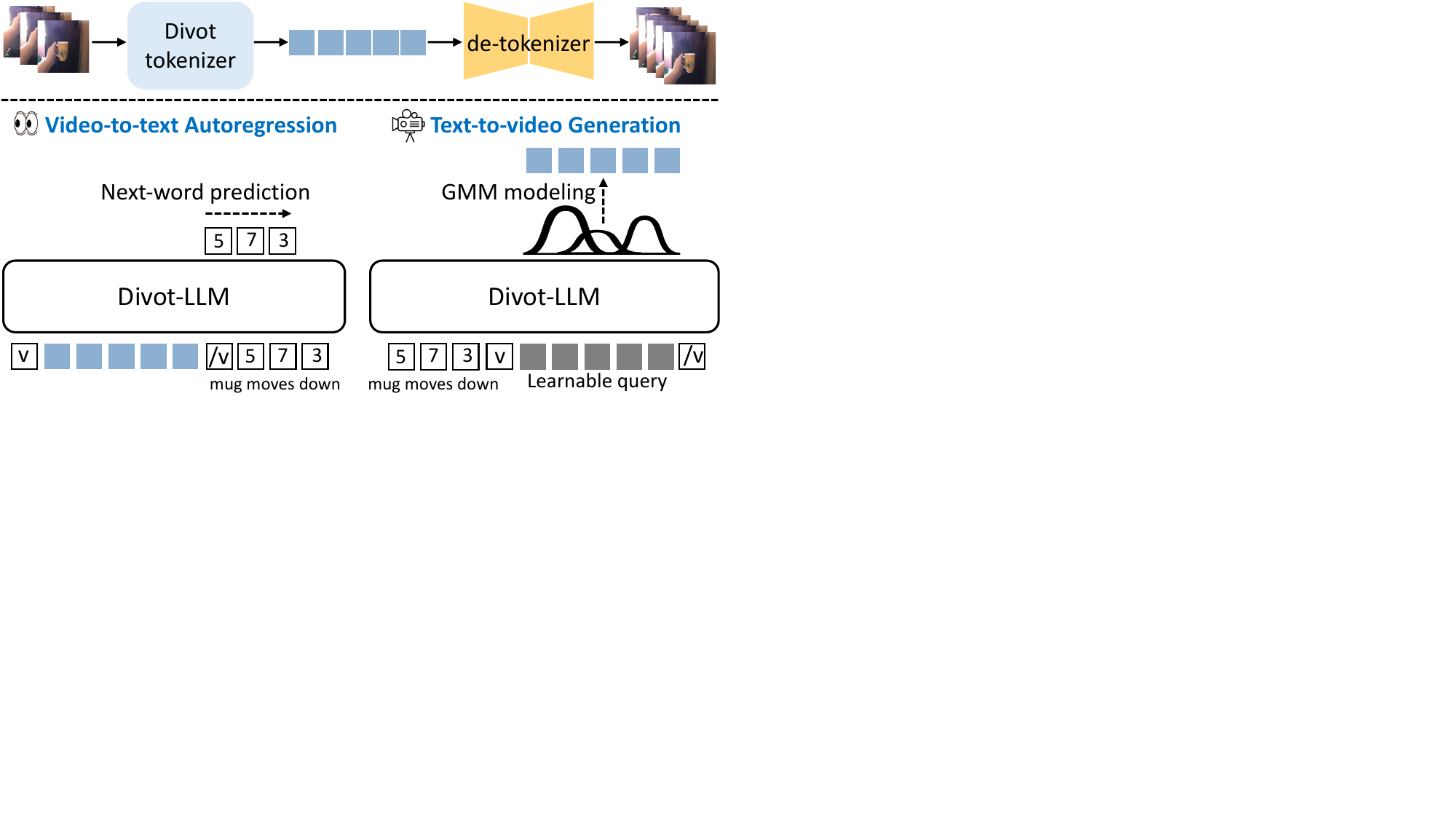}
	\caption{ Overview of \textbf{Divot-LLM}. Video features from the Divot tokenizer are fed into the LLM to perform next-word prediction for video comprehension, while learnable queries are input into the LLM to model the distributions of Divot features using a \textbf{Gaussian Mixture Model} (GMM) for video generation. During inference, video features are sampled from the predicted GMM distribution to decode videos using the de-tokenizer.}
	\label{fig:method_llm}
	\vspace{-4mm}	
\end{figure}

\begin{figure*}
	\centering
	\includegraphics[width=1.0\linewidth]{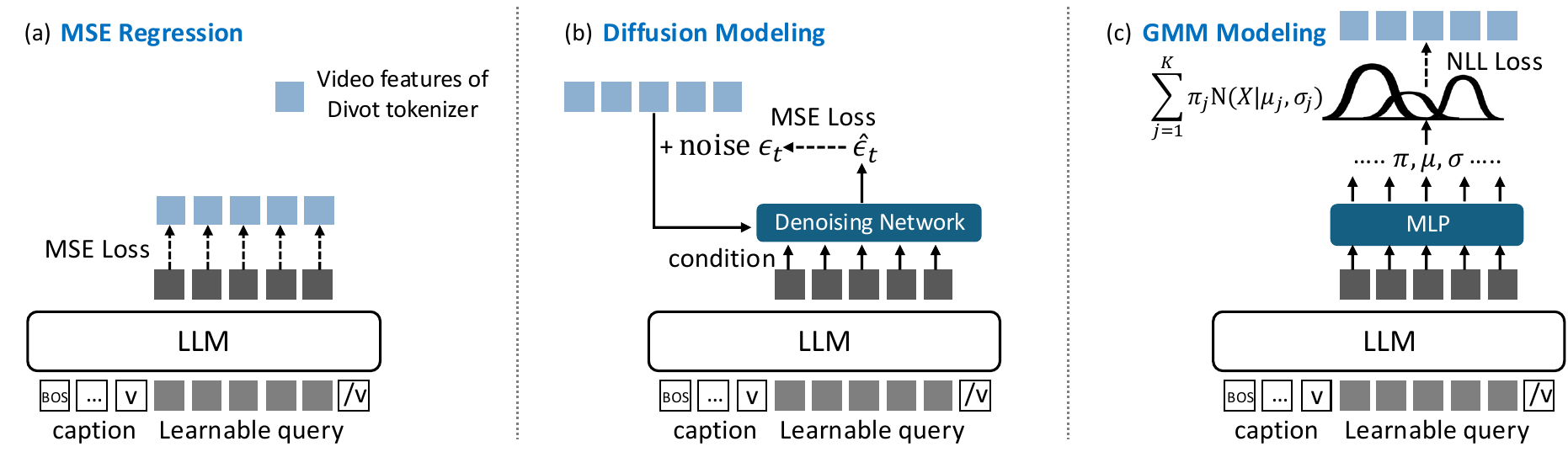}
	\caption{Paradigms for \textbf{modeling video representations} from the Divot tokenizer with a LLM for video generation. (a) \textbf{MSE Regression}, where the LLM output is trained to minimize its distance with video features using Mean Squared Error (MSE) loss; (b) \textbf{Diffusion Modeling}, where the LLM output is fed into a denoising network as the condition to predict the noise added to video features; (c) \textbf{GMM Modeling}, where the LLM output is trained to predict the parameters of a Gaussian Mixture Model (GMM) for modeling video feature distributions.}
	\label{fig:method_gen}
	\vspace{-1mm}	
\end{figure*}

\section{Method}
\label{sec:method}

\subsection{Divot Tokenizer}
We introduce Divot, a diffusion-powered tokenizer that leverage diffusion procedure for video representation learning. Additionally, the proxy diffusion model used for training the tokenizer can serve as a de-tokenizer to decode realistic video clips from their spatiotemporal representations. 

\subsubsection{Preliminary: Video Diffusion Model.} Diffusion models~\citep{ho2020denoising, rombach2022high} learns to model a probability distribution by reversing a process that progressively adds noise to the data. Specifically, given data $\mathbf{x}_0 \sim p(\mathbf{x})$, the forward process gradually adds random Gaussian noise $\mathbf{\epsilon_t} \in \mathcal{N}(\mathbf{0}, \mathbf{I})$ to the data sample $\mathbf{x}_0$ with a total of $T$ timesteps to yield $\mathbf{x}_t$ through a parameterization trick. The denoising process predicts $\mathbf{\epsilon_t}$ in the forward diffusion process with a denoising network $\mathbf{\epsilon}_{\theta}\left(\mathbf{x}_t, t\right)$, which is trained by the objective below,

\begin{equation}
\label{equ1}
\min _{\theta} \mathbb{E}_{t, \mathbf{x} \sim p, \mathbf{\epsilon} \sim \mathcal{N}(\mathbf{0}, \mathbf{I})}\Vert\mathbf{\epsilon}-\mathbf{\epsilon}_{\theta}\left(\mathbf{x}_t, t\right)\Vert_2^2,
\end{equation} 
where $\mathbf{\epsilon}$ is the sampled Gaussian noise and $\theta$ indicates the parameters of the denoising network. During inference, we can perform iterative denoising from a random Gaussian noise for the denoised data $\mathbf{x}_0$

For video diffusion models~\cite{chen2023videocrafter1, xing2025dynamicrafter}, given a video $\mathbf{x}$, a latent representation $\mathbf{z}=\mathcal{E}(\mathbf{x})$ is first encoded to reduce the computational complexity. Then the forward diffusion process and backward denoise process are performed in this latent space with a denoising network $\mathbf{\epsilon}_{\theta}\left(\mathbf{x}_t, \mathbf{c}, t\right)$, where $\mathbf{c}$ denotes denoising conditions like text or visual prompts.

\subsubsection{Training Pipeline}
As illustrated in Fig.~\ref{fig:method}, given a video clip, we separately sample sparse frames at 2 fps to obtain the video representations from the tokenizer, and sample dense frames at 8 fps to obtain latent representations $z_0$ from the frozen VAE~\cite{kingma2013auto} encoder. Sparse frames are sampled as the input of the video tokenizer considering the semantic redundancy between adjacent frames.
The forward diffusion process gradually adds Gaussian noise $\theta$ to $z_0$ for producing the noisy input $z_t$. At each backward step $t$, a denoising U-Net is trained to predict the noise added from the previous step to the current step by taking the time embedding and video representations as the condition. Specifically, the video representations interact with the denoising U-Net intermediate features through cross-attention layers, where each noisy latent attends to all video tokens. By constraining the U-Net to reconstruct fine-grained spatial and temporal information of video clips through relying on video features, the Divot tokenizer is optimized to capture both spatial characteristics and temporal dynamics for robust video representations. 
The Divot tokenizer is trained on pure videos of a subset of WebVid-10M~\cite{bain2021frozen} and Panda-70M~\cite{chen2024panda}, totaling 10M videos.

\begin{figure*}
	\centering
	\includegraphics[width=1.0\linewidth]{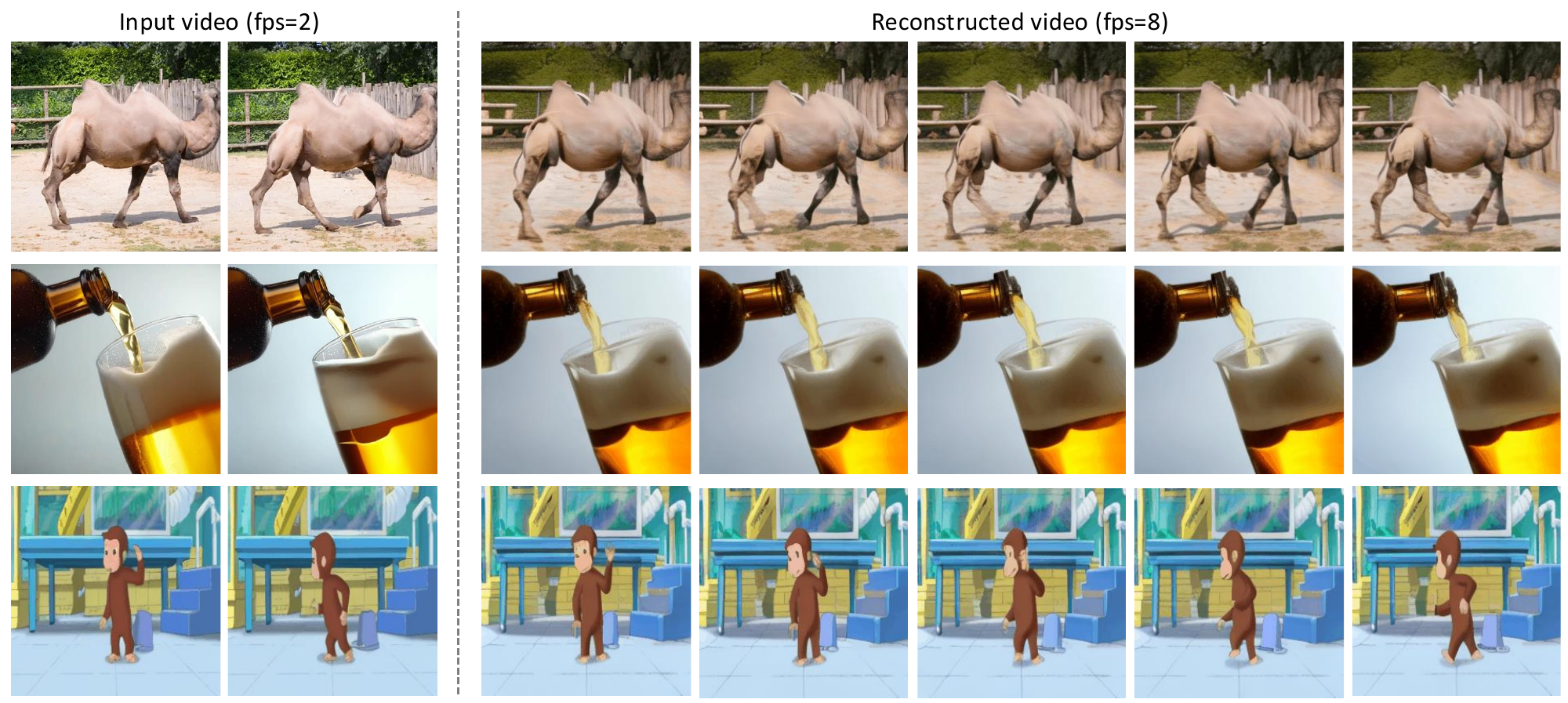}
	\caption{Reconstructed videos, where the Divot tokenizer obtains spatiotemporal representations of sparsely sampled video frames and the de-tokenizer decodes these representations into semantically aligned and temporally coherent video clips.}
	\label{fig:reconstruction}
	\vspace{-3mm}	
\end{figure*}

After training the Divot tokenizer, the proxy denoising U-Net (employed to implement the parameterized loss function) can serve as an effective video de-tokenizer, which is able to decode semantically aligned video clips from the learned spatiotemporal representations as shown in Fig.~\ref{fig:reconstruction}.

\subsubsection{Model Architecture}
As shown in Fig.~\ref{fig:method}, the Divot tokenizer is composed of a pre-trained ViT encoder to extract frame-level features, a transformer for spatial and temporal fusion, and a Perceiver Resampler~\cite{alayrac2022flamingo} to produce a fixed number of video tokens. The Perceiver Resampler is adopted for two reasons: (1) to reduce the number of video tokens that a LLM need to predict for generation, and (2) to transform the patch-position dependent features into a sequence of high-level features without 3D positional dependencies, which we empirically find easier for an LLM to fit (See Sec.~\ref{sec:ablation}). Specifically, given a video clip with a duration of two seconds, we sample 5 frames at 2 fps, resulting in a total of 64 video tokens. We adopt the de-noising U-Net in DynamiCrafter~\cite{xing2025dynamicrafter}, but reduce the input channel of the 3D convolution from 8 to 4 since we remove the original concatenation of a conditional image with noisy latents.

\subsection{Video Representation Modeling with LLM}
The core challenge of generating videos using a LLM with the Divot tokenizer lies in effectively modeling the continuous video features. The most straightforward solution is to minimize the distance between the LLM output and the video representations using mean squared error (MSE) loss following previous work~\cite{ge2024seed, sun2023emu2} for image generation as illustrated in Fig.~\ref{fig:method_gen} (a).
However, we empirically find that this approach is not effective for modeling continuous video features, as the generated videos tend to exhibit repeating patterns. We analyze that the deterministic regression regularizes the LLM to learn overly averaged representations, which is particularly catastrophic in video generation as videos must ensure both spatial and temporal diversity.

Inspired by recent work MAR~\cite{li2024autoregressive}, instead of deterministic regression, we aim to model the probability distribution of video representations using the LLM. As shown in Fig.~\ref{fig:method_gen}, we explore two approaches including (b) Diffusion Modeling~\cite{li2024autoregressive} and (c) GMM Modeling~\cite{tschannen2025givt}. Specifically, for the diffusion modeling, given continuous-valued video features to be predicted, the LLM produces output, which serves as the condition of a denoising network (a small MLP) to predict the Gaussian noise added to the video features. The diffusion model is trained for representing the distribution of video features. For GMM modeling, we use a Gaussian Mixture Model (GMM) to model the distribution of the video features, and train the LLM to predict $2kd + k$ parameters per video token ($kd$ mean and $kd$ variance parameters for the mixture components, and $k$ mixture probabilities). We optimize the LLM by minimizing the discrepancy between the predicted GMM distribution and the video representations with negative log-likelihood (NLL) loss.

During inference, in diffusion modeling, the denoising network denoise the final video features from Gaussian noise gradually by taking the LLM output as the condition. In GMM modeling, we draw samples from the predicted GMM distribution as the final video representations. To empirically investigate the effectiveness of the approaches above, we train the LLM with MSR-VTT~\cite{xu2016msr}, and evaluates text-to-video generation on test set with FVD~\cite{unterthiner2018towards} and similarity score~\cite{radford2021learning} as the metric following previous work~\cite{yan2021videogpt, jin2024video}. As listed in Tab.~\ref{tab:ablation_gen}, GMM modeling achieves better performance than diffusion modeling and MSE regression in video generation. We speculate that high-level features obtained by the Divot tokenizer are more sensitive to Gaussian noise compared to the VAE latents used by MAR, making training more challenging and resulting in suboptimal results. Therefore, we adopt GMM modeling to train Divot-LLM.

\begin{table}
\centering
\refstepcounter{table}
\caption{Datasets used for training the tokenizer and Divot-LLM.}
\label{tab:data}
\resizebox{1.0\columnwidth}{!}{
\begin{tblr}{
  row{4} = {c},
  row{6} = {c,m},
  row{7} = {c,m},
  row{8} = {c},
  row{9} = {c},
  cell{1}{2} = {c},
  cell{1}{3} = {c},
  cell{2}{2} = {c},
  cell{2}{3} = {c},
  cell{3}{1} = {r=2}{},
  cell{3}{2} = {c},
  cell{3}{3} = {c},
  cell{5}{1} = {r=5}{},
  cell{5}{2} = {c},
  cell{5}{3} = {c},
  vline{2-3} = {1-9}{},
  vline{3} = {4,6-9}{},
  hline{1,10} = {-}{0.08em},
  hline{2} = {-}{0.05em},
  hline{3,5} = {-}{},
  hline{4,6-9} = {2-3}{dashed},
}
\textbf{Stage}     & \textbf{Type}      & \textbf{Dataset}                                                           \\
Tokenize        & Pure Video         & WebVid-10M~\cite{bain2021frozen}, Panda-70M~\cite{chen2024panda}                                                      \\
Pre-train      & Video-text         & WebVid-10M~\cite{bain2021frozen}                                                                 \\
                   & Image-text         & CC3M~\cite{sharma2018conceptual},  CapsFusion~\cite{yu2024capsfusion}, LAION-COCO~\cite{schuhmann2022laion}                                             \\
SFT & Classification     & Kinetics-710~\cite{kay2017kinetics}, SSV2~\cite{goyal2017something}                                                         \\
                   & VQA                & {TGIF~\cite{li2016tgif}, NextQA~\cite{xiao2021next}, CLEVRER~\cite{yi2019clevrer},\\ YouCook2~\cite{zhou2018towards}, 
                   PerceptionTest\cite{patraucean2024perception},\\
                   EgoQA~\cite{grauman2022ego4d},
                    ActivityNetQA\cite{yu2019activitynet}} \\
                   & Instruction        & {Video-ChatGPT\cite{maaz2023video}, LLaVA-mixed\cite{liu2024visual}, \\ Valley~\cite{luo2023valley}, LLaVA-Video-178K\cite{lin2023video}}                  \\
                   & Generation   & WebVid-10M~\cite{bain2021frozen}                                                                 \\
                   & StoryTelling & In-house data                                                              
\end{tblr}}
\end{table}

\begin{table*}[t]
\centering
\small
\setlength\tabcolsep{5pt}
\caption{Comparison for video comprehension with MLLMs. ``Video-Gen'' denotes whether the model can generate videos besides texts. The evaluation metric is accuracy. The best results are bold and the second best results are underlined.}
\label{tab:video_comp}
\resizebox{1.8\columnwidth}{!}{
\begin{tabular}{lccccccc}
\toprule
\textbf{Model}            & \textbf{LLM size} & \textbf{Video-Gen} & \textbf{EgoSchema}     & \textbf{Perception-Test} & \textbf{MVBench}       & \textbf{MSVD} & \textbf{ActivityNet} \\
\midrule
Gemini 1.0 Pro~\cite{team2023gemini}   & -        & $\times$          & 55.7          & 51.1            & -             & -    & 49.8        \\
Gemini 1.5 Pro~\cite{team2024gemini}   & -        &$\times$            & 63.2          & -               & -             & -    & 56.7        \\
GPT4-V~\cite{openai2023gpt4v}           & -        &  $\times$          & 55.6          & -               & 43.7          & -    & 59.5        \\
GPT4-O~\cite{openai2024gpt4o}           & -        &  $\times$          & \textbf{72.2}          & -               & -             & -    & \underline{61.9}        \\
\midrule
LLaMA-VID~\cite{li2023llama}        & 7B       &  $\times$          & 38.5          & 44.6            & 41.9          & 69.7 & 47.4        \\
Video-ChatGPT~\cite{maaz2023video}  & 7B       &  $\times$          & -         & -            & -          & 64.9 & 35.2        \\
Video-LLaVA~\cite{lin2023video}      & 7B       &  $\times$          & 38.4          & 44.3            & 41.0          & 70.7 & 45.3        \\
VideoChat2~\cite{li2023mvbench}       & 7B       &  $\times$          & 42.2          & 47.3            & 51.1          & 70.0 & 49.1        \\
LLaVA-NeXT-Video~\cite{liu2024llavanext} & 7B       &   $\times$         & 43.9          & 48.8            & 46.5          & 67.8 & 53.5        \\
LLaVA-NeXT-Video~\cite{liu2024llavanext} & 32B      &   $\times$         & 60.9          & -               & -             & -    & 54.3        \\
PLLaVA~\cite{xu2024pllava}           & 34B      &   $\times$         & -             & 58.1   & -             & -    & 60.9        \\
LLaVA-OneVision~\cite{li2024llavaonevision}  & 72B      &  $\times$          & 62.0          & -               & -             & -    & \textbf{62.3}        \\
VideoLLaMA2~\cite{cheng2024videollama}      & 7B       & $\times$           & 51.7          & 51.4            & \underline{54.6}          & 70.9 & 50.2        \\
VideoLLaMA2~\cite{cheng2024videollama}      & 72B      & $\times$           & \underline{63.9} & \underline{57.5}            & \textbf{62.0} & 71.0 & 55.2        \\
LWM~\cite{liu2024world}     & 7B       & \checkmark          & -          & -            & -             & 55.9 & -        \\
Video-LaVIT~\cite{jin2024video}      & 7B       & \checkmark          & 37.3          & 47.9            & -             & 73.2 & 50.1        \\
VILA-U~\cite{wu2024vila}      & 7B       & \checkmark          & -          & -            & -             & \underline{75.3} & 52.7        \\
\midrule     
Divot-LLM     & 7B       & \checkmark          & 46.5          & \textbf{58.3}            & 52.1          &\textbf{76.4}      &55.8  \\ \bottomrule   
\end{tabular}}
\end{table*}

\subsection{Pre-training and Instruction Tuning}

\subsubsection{Training Stage I: Multimodal Pre-training}
As shown in Fig~\ref{fig:method_llm}, Divot-LLM adopts next-word prediction and GMM modeling on video-text data for video comprehension and generation. Specifically, the video features from the Divot tokenizer, the special tokens indicating the start and end of video features, along with the text tokens of the caption are fed into the pre-trained Mistral-7B~\cite{jiang2023mistral} for next token prediction trained with cross-entropy loss. Text tokens of the caption and $N$ learnable queries are input into the LLM, where the output of the learnable queries are trained via bidirectional attention to model a GMM distribution for the video features using NLL loss. During inference, we draw samples from the predicted GMM distribution as the condition of the denoising U-Net to decode realistic videos. We pre-train Divot-LLM from the pre-trained Mistral-7B model using LoRA~\cite{hu2021lora} on a subset of WebVid-10M~\cite{bain2021frozen} data (filtered for temporal dynamics in captions) and image-text data, utilizing 32 A100-40G GPUs.

\subsubsection{Training Stage II: Multimodal Instruction Tuning}
We perform multimodal instruction tuning on Divot-LLM to align it with human instructions
through supervised fine-tuning on public datasets as listed in Tab.~\ref{tab:data} with a LoRA module. 
We further fine-tune the pretrained Divot-LLM on an animated series called ``Curious George'' to achieve video storytelling, which generates storyline and corresponding video clips in an interleaved  manner.

\section{Experiment}

\subsection{Quantitative Evaluation}
{\flushleft \bf Video Comprehension.} We conduct extensive evaluations on video comprehension benchmarks including \textit{Multi-choice Video Question Answering} (MC-VQA) on EgoSchema~\cite{mangalam2023egoschema}, Perception-Test~\cite{patraucean2024perception}, MVBench~\cite{li2024mvbench}, and \textit{Open-Ended Video Question Answering} (OE-VQA) on MSVD~\cite{chen2011collecting}, ActivityNet~\cite{yu2019activitynet}. Following VideoLLaMA 2~\cite{cheng2024videollama}, we utilize GPT-3.5 to assess the quality of the generated answers of OE-VQA by determining whether the answers match the ground truth, and we report the percentage of ``Yes'' as Accuracy.

For each testing video, we sample a maximum of 20 clips, each containing 5 frames. The evaluation results are reported in Tab.~\ref{tab:video_comp}. Divot-LLM outperforms the baseline models that can generate both texts and videos, which demonstrates that our model effectively achieves video comprehension within a unified framework. Compared to VideoLLMs specifically designed for video comprehension of the same model size of LLM, Divot-LLM achieves competitive results with significantly fewer video-caption pairs for training (4.8M vs. 100M in VideoLLaMA 2). By utilizing diffusion procedure for video representation learning, our Divot tokenizer effectively captures robust spatiotemporal representations, enhancing the comprehension capabilities.

\begin{figure*}
	\centering
	\includegraphics[width=1.0\linewidth]{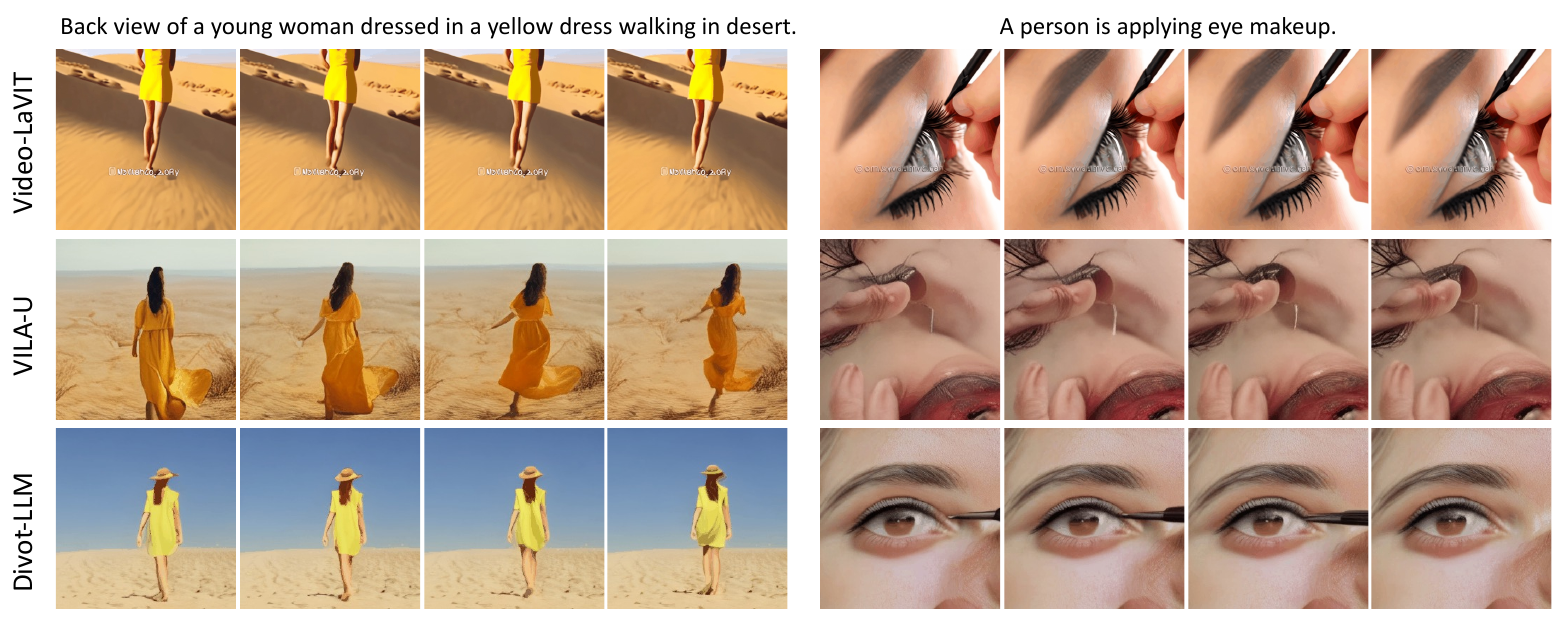}
	\caption{Qualitative comparison of text-to-video generation with MLLMs that are capable of unified video comprehension and generation. Divot-LLM effectively generates videos that are semantically aligned with text prompts, accurately reflecting temporal changes.}
	\label{fig:video_gen_comparison}
	\vspace{-2mm}	
\end{figure*}

{\flushleft \bf Video Generation.} We evaluate zero-shot text-to-video generation on MSR-VTT~\cite{xu2016msr}.
We randomly sample one caption for each testing video and generate 16 frames in 256 x 256px resolution. 
We adopt the CLIP similarity (CLIPSIM)~\cite{wu2021godiva} and Fr\'{e}chet video distance (FVD)~\cite{unterthiner2018towards} as the evaluation metric following Loong~\cite{wang2024loong}. As listed in Tab.~\ref{tab:t2v}, Divot-LLM achieves performance comparable to existing video generation models in terms of visual quality and semantic alignment with captions using only 4.8 million video-text pairs for training.

\subsection{Qualitative Evaluation}
{\flushleft \bf Text-to-video Generation.}
We perform qualitative comparison of text-to-video generation with baseline MLLMs that are capable of 
 unified video comprehension and generation. As illustrated in Fig.~\ref{fig:video_gen_comparison}, through modeling the distributions of Divot features with a predicted GMM, our Divot-LLM can generate videos that are both semantically aligned with text prompts and temporally coherent within frames. 

\begin{figure*}
	\centering
	\includegraphics[width=1.0\linewidth]{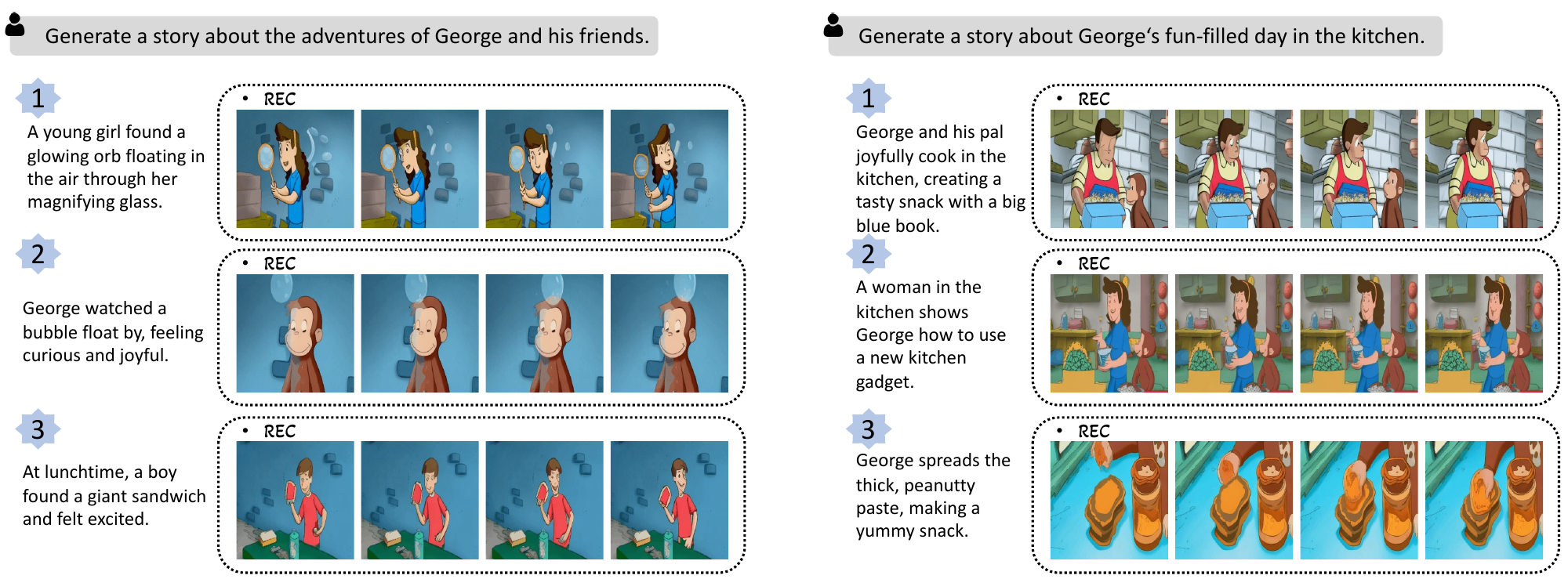}
	\caption{Qualitative examples of video storytelling by Divot-LLM. Given a story instruction, Divot-LLM can generate rich textual narratives along with corresponding video clips that are temporally coherent in an interleaved manner.}
	\label{fig:george_story}
\end{figure*}

{\flushleft \bf Video StoryTelling.} We fine-tune the pre-trained Divot-LLM on an animated series called ``Curious George'' for video storytelling. As shown in Fig.~\ref{fig:george_story}, given a brief story instruction, our Divot-LLM can generate a sequence of multimodal stories with rich narrative text and contextually relevant videos that are temporally coherent. Since we only fine-tune the de-tokenizer for adaptation to the new domain, it demonstrates the generalizability of our Divot tokenizer for obtaining robust video representations.

\begin{table}[!ht]
\caption{Comparison for zero-shot text-to-video generation. ``Data size'' refers to the number of training video data, and ``Unified'' denotes if the model enables video comprehension and generation. The best results are bold and the second best results are underlined.}
\vspace{-10pt}
\label{tab:t2v}
\begin{center}
\begin{small}
\resizebox{1.0\columnwidth}{!}{
\begin{tabular}{lcccc}
\toprule
\multirow{2}{*}{\textbf{Model}} & \multirow{2}{*}{\textbf{Data size}} & \multirow{2}{*}{\textbf{Unified}} & \multicolumn{2}{c}{\textbf{MSR-VTT}}\\
\cmidrule(lr){4-5}  & & & \textbf{CLIPSIM ($\uparrow$)} & \textbf{FVD ($\downarrow$)}\\
\midrule
CogVideo~\cite{hong2022cogvideo} & 5.4M & $\times$ & 0.2631 & 1294   \\
Video LDM~\cite{blattmann2023align} & 10M & $\times$ & 0.2929 & -   \\
VideoComposer~\cite{wang2024videocomposer} & 10M & $\times$ & 0.2932 & 580   \\
InternVid~\cite{wang2023internvid} & 28M & $\times$ & 0.2951 & -  \\
Make-A-Video~\cite{singer2022make} & 20M & $\times$ & \textbf{0.3049} & -  \\
VideoPoet~\cite{kondratyuk2023videopoet} & 270M & $\times$ & \textbf{0.3049} & 213  \\
PYoCo~\cite{ge2023preserve} & 22.5M & $\times$ & - & -   \\
SVD~\cite{blattmann2023stable} & 152M & $\times$ & - & -   \\
 Video-LavIT~\cite{jin2024video} & 10M & \checkmark & \underline{0.3012} & \underline{188.36}  \\
Loong~\cite{wang2024loong} &16M &$\times$ &0.2903 &274 \\
Snap Video~\cite{menapace2024snap}&- & $\times$ &0.2793 &\textbf{110.4} \\
 VILA-U~\cite{wu2024vila} & 1M & \checkmark &0.2937   &499.06  \\
\cmidrule{1-5} Divot-LLM & 4.8M & \checkmark &0.2938   &301.4  \\
\bottomrule
\end{tabular}}
\end{small}
\end{center}
\vspace{-10pt}
\end{table}

\subsection{Ablation Study}
\label{sec:ablation}

{\flushleft \bf Diffusion for Video Comprehension.} We design two baselines to validate the effectiveness of the diffusion procedure to learn spatiotemporal representations for VideoLLMs. As shown in Tab.~\ref{tab:ablation_comp}, both models are pre-trained on Valley~\cite{luo2023valley} and instruction tuned on Video-ChatGPT~\cite{maaz2023video}. The model with diffusion loss employs our Divot tokenizer, while the model with caption loss adopts the same architecture but its tokenizer is pre-trained using captioning loss with a frozen LLM on Valley. The model that employs a video tokenizer trained with diffusion loss achieves better performance in video comprehension benchmarks, demonstrating that the diffusion process can effectively learn robust video representations in a self-supervised manner, without the need for paired caption annotations.

\begin{table}
\centering
\refstepcounter{table}
\label{tab:ablation_comp}
\caption{Ablation study on the training objective of the video tokenizer. The evaluation metric is accuracy.}
\resizebox{0.8\columnwidth}{!}{
\begin{tblr}{
  cell{1-3}{2} = {c},
  cell{1-3}{3} = {c},
  cell{1-3}{4} = {c},
  vline{2} = {-}{},
  hline{2} = {-}{},
  hline{1,4} = {-}{0.08em},
}
\textbf{Loss Type} & \textbf{MV-Bench} & \textbf{MSVD} & \textbf{ActivityNet} \\

Caption   & 30.8       &   66.1             & 43.2 \\ 
Diffusion & \textbf{33.2}         & \textbf{68.9}                & \textbf{44.3}             \\
\end{tblr}}
\vspace{-5pt}
\end{table}

\begin{table*}
\centering
\refstepcounter{table}
\label{tab:ablation_gen}
\caption{Ablation study on video representation modeling with LLMs for generation. We evaluate text-to-video generation on MSR-VTT.}
\resizebox{1.8\columnwidth}{!}{
\begin{tblr}{
  column{even} = {c},
  column{3} = {c},
  column{5} = {c},
  column{9} = {c},
  cell{1}{2} = {c=2}{},
  cell{1}{4} = {c=4}{},
  cell{1}{8} = {c=3}{},
  cell{2}{2} = {r=2}{},
  cell{2}{3} = {r=2}{},
  cell{2}{5} = {c=2}{},
  cell{2}{7} = {c},
  cell{2}{9} = {c=2}{},
  cell{4}{7} = {c},
  cell{5}{7} = {c},
  vline{2,4,8} = {-}{0.05em}, 
  hline{1,6} = {-}{0.08em},
  hline{2} = {2-10}{0.03em},
  hline{3} = {2-10}{dashed},
  hline{4} = {-}{0.05em},
}
        & \textbf{\textbf{Representation}} &                               & \textbf{Objective } &                     &        &                 & \textbf{Mechanism} &         &                 \\
        & {patch-position\\dependent }     & {patch-position\\independent } & MSE            & Diffusion           &        & GMM             & AR                 & Query   &                 \\
        &                                  &                               &                & $\mathbf{\epsilon}$-pred & $v$-pred    &                 &                    & ~causal & bidirectional   \\
CLIPSIM ($\uparrow$) & 0.3192                           & \textbf{0.3265}               & 0.3168         & 0.2811              & 0.2842 & \textbf{0.3265} & 0.2386             & 0.3080  & \textbf{0.3265} \\
FVD ($\downarrow$)    & 378.50                           & \textbf{366.60}               & 438.94         & 418.19              & 377.17 & \textbf{366.60} & 447.88             & 416.60  & \textbf{366.60} 
\end{tblr}}
\vspace{-2mm}
\end{table*}

{\flushleft \bf Video Generation with LLM.} 
We perform various ablation studies to explore an effective approach for generating videos with a LLM through training on MSR-VTT training set and evaluating text-to-video generation on test set. We use ViT-G/14 to calculate CLIPSIM for better discrimination. 

\textbf{Q1: Which type of video representations is easier?} We investigate two types of video representations, patch-position dependent features obtained from a spatial-temporal transformer and patch-position independent features after a Perceiver Resample with learnable queries. As listed in Tab.~\ref{tab:ablation_gen}, fitting features without 3D positional dependencies achieves higher performance, which is also observed in recent work~\cite{wang2024larp}. We also experiment with training the video tokenizer in the VAE manner, which involves predicting the means and variances of a normal distribution and sampling video representations using the re-parametrization trick following GIVT~\cite{tschannen2025givt}. However, we observe that it is difficult for the LLM to converge and the video de-tokenizer achieves unsatisfactory reconstruction results. We conclude that \textit{introducing variances during tokenization for high-level video features may not be appropriate}.

\textbf{Q2: Which training objective is more suitable?} As introduce in Sec.~\ref{sec:method}, we explore MSE regression, Diffusion modeling and GMM modeling to fit high-level continuous features with a LLM. As listed in Tab.~\ref{tab:ablation_gen}, simply aligning the LLM output with video features using MSE loss yields the lowest generation quality, suggesting that deterministic regression is inadequate for modeling spatiotemporal representations. Training a denoising network to denoise the noisy video features by taking the LLM output as the condition also achieves inferior performance with both $\mathbf{\epsilon}$ prediction and $v$ prediction. Different from MAR~\cite{li2024autoregressive} that denoises low-level VAE latents, our goal is to denoise high-level video features. We speculate that these features are more sensitive to Gaussian noise, making them more challenging to denoise. Training the LLM to model the distribution of high-level video features using a GMM model achieves the best generation quality and semantic alignment with captions. 

\textbf{Q3: Which LLM mechanism is more effective?} We train the LLM to fit the video representations with GMM modeling using both an autoregressive approach and a query-based approach, with the latter exploring causal attention and bidirectional attention within the LLM. Predicting the features of each video token in an autoregressive manner results in the worst performance due to the accumulation of errors, particularly when the features of the pervious token are sampled from a GMM distribution for predicting the distribution of the current token. The query-based approach achieves better results with bidirectional attention, as it enables each query to attend to all tokens for predictions.

\section{Conclusion}
In this work, we introduce Divot, a diffusion-powered video tokenizer learned in a self-supervised manner for unified comprehension and generation. We further investigate effective approaches for modeling continuous video representations with the LLM and present Divot-LLM to understand and generate video content in a single framework. Divot-LLM achieves competitive performance in video comprehension and generation benchmarks, and enables video storytelling effectively. We hope our work will draw increased attention to unifying video comprehension and generation through the design of sophisticated tokenizers.

{\flushleft \bf Limitation.} 
As we primarily focus on exploring effective representations and approaches for video generation with a unified LLM, the current model is trained to predict video representations for only a single clip and does not generate longer videos, which will be explored in our future work.

{
    \small
    \bibliographystyle{ieeenat_fullname}
    \bibliography{main}

\begin{thebibliography}{93}
\providecommand{\natexlab}[1]{#1}
\providecommand{\url}[1]{\texttt{#1}}
\expandafter\ifx\csname urlstyle\endcsname\relax
  \providecommand{\doi}[1]{doi: #1}\else
  \providecommand{\doi}{doi: \begingroup \urlstyle{rm}\Url}\fi

\bibitem[Alayrac et~al.(2022)Alayrac, Donahue, Luc, Miech, Barr, Hasson, Lenc, Mensch, Millican, Reynolds, et~al.]{alayrac2022flamingo}
Jean-Baptiste Alayrac, Jeff Donahue, Pauline Luc, Antoine Miech, Iain Barr, Yana Hasson, Karel Lenc, Arthur Mensch, Katherine Millican, Malcolm Reynolds, et~al.
\newblock Flamingo: a visual language model for few-shot learning.
\newblock \emph{Advances in Neural Information Processing Systems}, 35:\penalty0 23716--23736, 2022.

\bibitem[Bain et~al.(2021)Bain, Nagrani, Varol, and Zisserman]{bain2021frozen}
Max Bain, Arsha Nagrani, G{\"u}l Varol, and Andrew Zisserman.
\newblock Frozen in time: A joint video and image encoder for end-to-end retrieval.
\newblock In \emph{Proceedings of the IEEE/CVF international conference on computer vision}, pages 1728--1738, 2021.

\bibitem[Baranchuk et~al.(2021)Baranchuk, Rubachev, Voynov, Khrulkov, and Babenko]{baranchuk2021label}
Dmitry Baranchuk, Ivan Rubachev, Andrey Voynov, Valentin Khrulkov, and Artem Babenko.
\newblock Label-efficient semantic segmentation with diffusion models.
\newblock \emph{arXiv preprint arXiv:2112.03126}, 2021.

\bibitem[Blattmann et~al.(2023{\natexlab{a}})Blattmann, Dockhorn, Kulal, Mendelevitch, Kilian, Lorenz, Levi, English, Voleti, Letts, et~al.]{blattmann2023stable}
Andreas Blattmann, Tim Dockhorn, Sumith Kulal, Daniel Mendelevitch, Maciej Kilian, Dominik Lorenz, Yam Levi, Zion English, Vikram Voleti, Adam Letts, et~al.
\newblock Stable video diffusion: Scaling latent video diffusion models to large datasets.
\newblock \emph{arXiv preprint arXiv:2311.15127}, 2023{\natexlab{a}}.

\bibitem[Blattmann et~al.(2023{\natexlab{b}})Blattmann, Rombach, Ling, Dockhorn, Kim, Fidler, and Kreis]{blattmann2023align}
Andreas Blattmann, Robin Rombach, Huan Ling, Tim Dockhorn, Seung~Wook Kim, Sanja Fidler, and Karsten Kreis.
\newblock Align your latents: High-resolution video synthesis with latent diffusion models.
\newblock In \emph{Proceedings of the IEEE/CVF Conference on Computer Vision and Pattern Recognition}, pages 22563--22575, 2023{\natexlab{b}}.

\bibitem[Brown et~al.(2020)Brown, Mann, Ryder, Subbiah, Kaplan, Dhariwal, Neelakantan, Shyam, Sastry, Askell, et~al.]{brown2020language}
Tom Brown, Benjamin Mann, Nick Ryder, Melanie Subbiah, Jared~D Kaplan, Prafulla Dhariwal, Arvind Neelakantan, Pranav Shyam, Girish Sastry, Amanda Askell, et~al.
\newblock Language models are few-shot learners.
\newblock \emph{Advances in neural information processing systems}, 33:\penalty0 1877--1901, 2020.

\bibitem[Chen and Dolan(2011)]{chen2011collecting}
David Chen and William~B Dolan.
\newblock Collecting highly parallel data for paraphrase evaluation.
\newblock In \emph{Proceedings of the 49th annual meeting of the association for computational linguistics: human language technologies}, pages 190--200, 2011.

\bibitem[Chen et~al.(2023)Chen, Xia, He, Zhang, Cun, Yang, Xing, Liu, Chen, Wang, et~al.]{chen2023videocrafter1}
Haoxin Chen, Menghan Xia, Yingqing He, Yong Zhang, Xiaodong Cun, Shaoshu Yang, Jinbo Xing, Yaofang Liu, Qifeng Chen, Xintao Wang, et~al.
\newblock Videocrafter1: Open diffusion models for high-quality video generation.
\newblock \emph{arXiv preprint arXiv:2310.19512}, 2023.

\bibitem[Chen et~al.(2024)Chen, Siarohin, Menapace, Deyneka, Chao, Jeon, Fang, Lee, Ren, Yang, et~al.]{chen2024panda}
Tsai-Shien Chen, Aliaksandr Siarohin, Willi Menapace, Ekaterina Deyneka, Hsiang-wei Chao, Byung~Eun Jeon, Yuwei Fang, Hsin-Ying Lee, Jian Ren, Ming-Hsuan Yang, et~al.
\newblock Panda-70m: Captioning 70m videos with multiple cross-modality teachers.
\newblock In \emph{Proceedings of the IEEE/CVF Conference on Computer Vision and Pattern Recognition}, pages 13320--13331, 2024.

\bibitem[Cheng et~al.(2024)Cheng, Leng, Zhang, Xin, Li, Chen, Zhu, Zhang, Luo, Zhao, et~al.]{cheng2024videollama}
Zesen Cheng, Sicong Leng, Hang Zhang, Yifei Xin, Xin Li, Guanzheng Chen, Yongxin Zhu, Wenqi Zhang, Ziyang Luo, Deli Zhao, et~al.
\newblock Videollama 2: Advancing spatial-temporal modeling and audio understanding in video-llms.
\newblock \emph{arXiv preprint arXiv:2406.07476}, 2024.

\bibitem[Chowdhery et~al.(2022)Chowdhery, Narang, Devlin, Bosma, Mishra, Roberts, Barham, Chung, Sutton, Gehrmann, et~al.]{chowdhery2022palm}
Aakanksha Chowdhery, Sharan Narang, Jacob Devlin, Maarten Bosma, Gaurav Mishra, Adam Roberts, Paul Barham, Hyung~Won Chung, Charles Sutton, Sebastian Gehrmann, et~al.
\newblock Palm: Scaling language modeling with pathways.
\newblock \emph{arXiv preprint arXiv:2204.02311}, 2022.

\bibitem[Dong et~al.(2023)Dong, Han, Peng, Qi, Ge, Yang, Zhao, Sun, Zhou, Wei, et~al.]{dong2023dreamllm}
Runpei Dong, Chunrui Han, Yuang Peng, Zekun Qi, Zheng Ge, Jinrong Yang, Liang Zhao, Jianjian Sun, Hongyu Zhou, Haoran Wei, et~al.
\newblock Dreamllm: Synergistic multimodal comprehension and creation.
\newblock \emph{arXiv preprint arXiv:2309.11499}, 2023.

\bibitem[Dosovitskiy et~al.(2020)Dosovitskiy, Beyer, Kolesnikov, Weissenborn, Zhai, Unterthiner, Dehghani, Minderer, Heigold, Gelly, et~al.]{dosovitskiy2020image}
Alexey Dosovitskiy, Lucas Beyer, Alexander Kolesnikov, Dirk Weissenborn, Xiaohua Zhai, Thomas Unterthiner, Mostafa Dehghani, Matthias Minderer, Georg Heigold, Sylvain Gelly, et~al.
\newblock An image is worth 16x16 words: Transformers for image recognition at scale.
\newblock \emph{arXiv preprint arXiv:2010.11929}, 2020.

\bibitem[Ge et~al.(2023{\natexlab{a}})Ge, Nah, Liu, Poon, Tao, Catanzaro, Jacobs, Huang, Liu, and Balaji]{ge2023preserve}
Songwei Ge, Seungjun Nah, Guilin Liu, Tyler Poon, Andrew Tao, Bryan Catanzaro, David Jacobs, Jia-Bin Huang, Ming-Yu Liu, and Yogesh Balaji.
\newblock Preserve your own correlation: A noise prior for video diffusion models.
\newblock In \emph{Proceedings of the IEEE/CVF International Conference on Computer Vision}, pages 22930--22941, 2023{\natexlab{a}}.

\bibitem[Ge et~al.(2023{\natexlab{b}})Ge, Ge, Zeng, Wang, and Shan]{ge2023planting}
Yuying Ge, Yixiao Ge, Ziyun Zeng, Xintao Wang, and Ying Shan.
\newblock Planting a seed of vision in large language model.
\newblock \emph{arXiv preprint arXiv:2307.08041}, 2023{\natexlab{b}}.

\bibitem[Ge et~al.(2023{\natexlab{c}})Ge, Zhao, Zeng, Ge, Li, Wang, and Shan]{ge2023making}
Yuying Ge, Sijie Zhao, Ziyun Zeng, Yixiao Ge, Chen Li, Xintao Wang, and Ying Shan.
\newblock Making llama see and draw with seed tokenizer.
\newblock \emph{arXiv preprint arXiv:2310.01218}, 2023{\natexlab{c}}.

\bibitem[Ge et~al.(2024)Ge, Zhao, Zhu, Ge, Yi, Song, Li, Ding, and Shan]{ge2024seed}
Yuying Ge, Sijie Zhao, Jinguo Zhu, Yixiao Ge, Kun Yi, Lin Song, Chen Li, Xiaohan Ding, and Ying Shan.
\newblock Seed-x: Multimodal models with unified multi-granularity comprehension and generation.
\newblock \emph{arXiv preprint arXiv:2404.14396}, 2024.

\bibitem[Goyal et~al.(2017)Goyal, Ebrahimi~Kahou, Michalski, Materzynska, Westphal, Kim, Haenel, Fruend, Yianilos, Mueller-Freitag, et~al.]{goyal2017something}
Raghav Goyal, Samira Ebrahimi~Kahou, Vincent Michalski, Joanna Materzynska, Susanne Westphal, Heuna Kim, Valentin Haenel, Ingo Fruend, Peter Yianilos, Moritz Mueller-Freitag, et~al.
\newblock The" something something" video database for learning and evaluating visual common sense.
\newblock In \emph{Proceedings of the IEEE international conference on computer vision}, pages 5842--5850, 2017.

\bibitem[Grauman et~al.(2022)Grauman, Westbury, Byrne, Chavis, Furnari, Girdhar, Hamburger, Jiang, Liu, Liu, et~al.]{grauman2022ego4d}
Kristen Grauman, Andrew Westbury, Eugene Byrne, Zachary Chavis, Antonino Furnari, Rohit Girdhar, Jackson Hamburger, Hao Jiang, Miao Liu, Xingyu Liu, et~al.
\newblock Ego4d: Around the world in 3,000 hours of egocentric video.
\newblock In \emph{Proceedings of the IEEE/CVF Conference on Computer Vision and Pattern Recognition}, pages 18995--19012, 2022.

\bibitem[Ho et~al.(2020)Ho, Jain, and Abbeel]{ho2020denoising}
Jonathan Ho, Ajay Jain, and Pieter Abbeel.
\newblock Denoising diffusion probabilistic models.
\newblock \emph{Advances in neural information processing systems}, 33:\penalty0 6840--6851, 2020.

\bibitem[Hong et~al.(2022)Hong, Ding, Zheng, Liu, and Tang]{hong2022cogvideo}
Wenyi Hong, Ming Ding, Wendi Zheng, Xinghan Liu, and Jie Tang.
\newblock Cogvideo: Large-scale pretraining for text-to-video generation via transformers.
\newblock \emph{arXiv preprint arXiv:2205.15868}, 2022.

\bibitem[Hu et~al.(2021)Hu, Shen, Wallis, Allen-Zhu, Li, Wang, Wang, and Chen]{hu2021lora}
Edward~J Hu, Yelong Shen, Phillip Wallis, Zeyuan Allen-Zhu, Yuanzhi Li, Shean Wang, Lu Wang, and Weizhu Chen.
\newblock Lora: Low-rank adaptation of large language models.
\newblock \emph{arXiv preprint arXiv:2106.09685}, 2021.

\bibitem[Hudson et~al.(2024)Hudson, Zoran, Malinowski, Lampinen, Jaegle, McClelland, Matthey, Hill, and Lerchner]{hudson2024soda}
Drew~A Hudson, Daniel Zoran, Mateusz Malinowski, Andrew~K Lampinen, Andrew Jaegle, James~L McClelland, Loic Matthey, Felix Hill, and Alexander Lerchner.
\newblock Soda: Bottleneck diffusion models for representation learning.
\newblock In \emph{Proceedings of the IEEE/CVF Conference on Computer Vision and Pattern Recognition}, pages 23115--23127, 2024.

\bibitem[Jiang et~al.(2023)Jiang, Sablayrolles, Mensch, Bamford, Chaplot, Casas, Bressand, Lengyel, Lample, Saulnier, et~al.]{jiang2023mistral}
Albert~Q Jiang, Alexandre Sablayrolles, Arthur Mensch, Chris Bamford, Devendra~Singh Chaplot, Diego de~las Casas, Florian Bressand, Gianna Lengyel, Guillaume Lample, Lucile Saulnier, et~al.
\newblock Mistral 7b.
\newblock \emph{arXiv preprint arXiv:2310.06825}, 2023.

\bibitem[Jin et~al.(2023)Jin, Xu, Chen, Liao, Tan, Chen, Lei, Liu, Song, Lei, et~al.]{jin2023unified}
Yang Jin, Kun Xu, Liwei Chen, Chao Liao, Jianchao Tan, Bin Chen, Chenyi Lei, An Liu, Chengru Song, Xiaoqiang Lei, et~al.
\newblock Unified language-vision pretraining with dynamic discrete visual tokenization.
\newblock \emph{arXiv preprint arXiv:2309.04669}, 2023.

\bibitem[Jin et~al.(2024)Jin, Sun, Xu, Chen, Jiang, Huang, Song, Liu, Zhang, Song, et~al.]{jin2024video}
Yang Jin, Zhicheng Sun, Kun Xu, Liwei Chen, Hao Jiang, Quzhe Huang, Chengru Song, Yuliang Liu, Di Zhang, Yang Song, et~al.
\newblock Video-lavit: Unified video-language pre-training with decoupled visual-motional tokenization.
\newblock \emph{arXiv preprint arXiv:2402.03161}, 2024.

\bibitem[Kay et~al.(2017)Kay, Carreira, Simonyan, Zhang, Hillier, Vijayanarasimhan, Viola, Green, Back, Natsev, et~al.]{kay2017kinetics}
Will Kay, Joao Carreira, Karen Simonyan, Brian Zhang, Chloe Hillier, Sudheendra Vijayanarasimhan, Fabio Viola, Tim Green, Trevor Back, Paul Natsev, et~al.
\newblock The kinetics human action video dataset.
\newblock \emph{arXiv preprint arXiv:1705.06950}, 2017.

\bibitem[Kingma and Welling(2013)]{kingma2013auto}
Diederik~P Kingma and Max Welling.
\newblock Auto-encoding variational bayes.
\newblock \emph{arXiv preprint arXiv:1312.6114}, 2013.

\bibitem[Kondratyuk et~al.(2023)Kondratyuk, Yu, Gu, Lezama, Huang, Schindler, Hornung, Birodkar, Yan, Chiu, et~al.]{kondratyuk2023videopoet}
Dan Kondratyuk, Lijun Yu, Xiuye Gu, Jos{\'e} Lezama, Jonathan Huang, Grant Schindler, Rachel Hornung, Vighnesh Birodkar, Jimmy Yan, Ming-Chang Chiu, et~al.
\newblock Videopoet: A large language model for zero-shot video generation.
\newblock \emph{arXiv preprint arXiv:2312.14125}, 2023.

\bibitem[Li et~al.(2024{\natexlab{a}})Li, Zhang, Guo, Zhang, Li, Zhang, Zhang, Li, Liu, and Li]{li2024llavaonevision}
Bo Li, Yuanhan Zhang, Dong Guo, Renrui Zhang, Feng Li, Hao Zhang, Kaichen Zhang, Yanwei Li, Ziwei Liu, and Chunyuan Li.
\newblock Llava-onevision: Easy visual task transfer.
\newblock \emph{arXiv preprint arXiv:2408.03326}, 2024{\natexlab{a}}.

\bibitem[Li et~al.(2023{\natexlab{a}})Li, Wang, He, Li, Wang, Liu, Wang, Xu, Chen, Luo, et~al.]{li2023mvbench}
Kunchang Li, Yali Wang, Yinan He, Yizhuo Li, Yi Wang, Yi Liu, Zun Wang, Jilan Xu, Guo Chen, Ping Luo, et~al.
\newblock Mvbench: A comprehensive multi-modal video understanding benchmark.
\newblock \emph{arXiv preprint arXiv:2311.17005}, 2023{\natexlab{a}}.

\bibitem[Li et~al.(2024{\natexlab{b}})Li, Wang, He, Li, Wang, Liu, Wang, Xu, Chen, Luo, et~al.]{li2024mvbench}
Kunchang Li, Yali Wang, Yinan He, Yizhuo Li, Yi Wang, Yi Liu, Zun Wang, Jilan Xu, Guo Chen, Ping Luo, et~al.
\newblock Mvbench: A comprehensive multi-modal video understanding benchmark.
\newblock In \emph{Proceedings of the IEEE/CVF Conference on Computer Vision and Pattern Recognition}, pages 22195--22206, 2024{\natexlab{b}}.

\bibitem[Li et~al.(2024{\natexlab{c}})Li, Tian, Li, Deng, and He]{li2024autoregressive}
Tianhong Li, Yonglong Tian, He Li, Mingyang Deng, and Kaiming He.
\newblock Autoregressive image generation without vector quantization.
\newblock \emph{arXiv preprint arXiv:2406.11838}, 2024{\natexlab{c}}.

\bibitem[Li et~al.(2016)Li, Song, Cao, Tetreault, Goldberg, Jaimes, and Luo]{li2016tgif}
Yuncheng Li, Yale Song, Liangliang Cao, Joel Tetreault, Larry Goldberg, Alejandro Jaimes, and Jiebo Luo.
\newblock Tgif: A new dataset and benchmark on animated gif description.
\newblock In \emph{Proceedings of the IEEE Conference on Computer Vision and Pattern Recognition}, pages 4641--4650, 2016.

\bibitem[Li et~al.(2023{\natexlab{b}})Li, Wang, and Jia]{li2023llama}
Yanwei Li, Chengyao Wang, and Jiaya Jia.
\newblock Llama-vid: An image is worth 2 tokens in large language models.
\newblock \emph{arXiv preprint arXiv:2311.17043}, 2023{\natexlab{b}}.

\bibitem[Li et~al.(2024{\natexlab{d}})Li, Zhang, Wang, Zhong, Chen, Chu, Liu, and Jia]{li2024mini}
Yanwei Li, Yuechen Zhang, Chengyao Wang, Zhisheng Zhong, Yixin Chen, Ruihang Chu, Shaoteng Liu, and Jiaya Jia.
\newblock Mini-gemini: Mining the potential of multi-modality vision language models.
\newblock \emph{arXiv preprint arXiv:2403.18814}, 2024{\natexlab{d}}.

\bibitem[Lin et~al.(2023)Lin, Zhu, Ye, Ning, Jin, and Yuan]{lin2023video}
Bin Lin, Bin Zhu, Yang Ye, Munan Ning, Peng Jin, and Li Yuan.
\newblock Video-llava: Learning united visual representation by alignment before projection.
\newblock \emph{arXiv preprint arXiv:2311.10122}, 2023.

\bibitem[Liu et~al.(2024{\natexlab{a}})Liu, Li, Li, Li, Zhang, Shen, and Lee]{liu2024llavanext}
Haotian Liu, Chunyuan Li, Yuheng Li, Bo Li, Yuanhan Zhang, Sheng Shen, and Yong~Jae Lee.
\newblock Llava-next: Improved reasoning, ocr, and world knowledge, 2024{\natexlab{a}}.

\bibitem[Liu et~al.(2024{\natexlab{b}})Liu, Li, Wu, and Lee]{liu2024visual}
Haotian Liu, Chunyuan Li, Qingyang Wu, and Yong~Jae Lee.
\newblock Visual instruction tuning.
\newblock \emph{Advances in neural information processing systems}, 36, 2024{\natexlab{b}}.

\bibitem[Liu et~al.(2024{\natexlab{c}})Liu, Yan, Zaharia, and Abbeel]{liu2024world}
Hao Liu, Wilson Yan, Matei Zaharia, and Pieter Abbeel.
\newblock World model on million-length video and language with ringattention.
\newblock \emph{arXiv preprint arXiv:2402.08268}, 2024{\natexlab{c}}.

\bibitem[Lu et~al.(2023)Lu, Clark, Lee, Zhang, Khosla, Marten, Hoiem, and Kembhavi]{lu2023unified}
Jiasen Lu, Christopher Clark, Sangho Lee, Zichen Zhang, Savya Khosla, Ryan Marten, Derek Hoiem, and Aniruddha Kembhavi.
\newblock Unified-io 2: Scaling autoregressive multimodal models with vision, language, audio, and action.
\newblock \emph{arXiv preprint arXiv:2312.17172}, 2023.

\bibitem[Luo et~al.(2023)Luo, Zhao, Yang, Dong, Li, Lu, Wang, Hu, Qiu, and Wei]{luo2023valley}
Ruipu Luo, Ziwang Zhao, Min Yang, Junwei Dong, Da Li, Pengcheng Lu, Tao Wang, Linmei Hu, Minghui Qiu, and Zhongyu Wei.
\newblock Valley: Video assistant with large language model enhanced ability.
\newblock \emph{arXiv preprint arXiv:2306.07207}, 2023.

\bibitem[Maaz et~al.(2023)Maaz, Rasheed, Khan, and Khan]{maaz2023video}
Muhammad Maaz, Hanoona Rasheed, Salman Khan, and Fahad~Shahbaz Khan.
\newblock Video-chatgpt: Towards detailed video understanding via large vision and language models.
\newblock \emph{arXiv preprint arXiv:2306.05424}, 2023.

\bibitem[Mangalam et~al.(2023)Mangalam, Akshulakov, and Malik]{mangalam2023egoschema}
Karttikeya Mangalam, Raiymbek Akshulakov, and Jitendra Malik.
\newblock Egoschema: A diagnostic benchmark for very long-form video language understanding.
\newblock \emph{Advances in Neural Information Processing Systems}, 36:\penalty0 46212--46244, 2023.

\bibitem[Menapace et~al.(2024)Menapace, Siarohin, Skorokhodov, Deyneka, Chen, Kag, Fang, Stoliar, Ricci, Ren, et~al.]{menapace2024snap}
Willi Menapace, Aliaksandr Siarohin, Ivan Skorokhodov, Ekaterina Deyneka, Tsai-Shien Chen, Anil Kag, Yuwei Fang, Aleksei Stoliar, Elisa Ricci, Jian Ren, et~al.
\newblock Snap video: Scaled spatiotemporal transformers for text-to-video synthesis.
\newblock In \emph{Proceedings of the IEEE/CVF Conference on Computer Vision and Pattern Recognition}, pages 7038--7048, 2024.

\bibitem[OpenAI(2023)]{openai2023gpt4v}
OpenAI.
\newblock Gpt-4v(ision) system card, 2023.

\bibitem[OpenAI(2024)]{openai2024gpt4o}
OpenAI.
\newblock Gpt-4o system card, 2024.

\bibitem[Patraucean et~al.(2024)Patraucean, Smaira, Gupta, Recasens, Markeeva, Banarse, Koppula, Malinowski, Yang, Doersch, et~al.]{patraucean2024perception}
Viorica Patraucean, Lucas Smaira, Ankush Gupta, Adria Recasens, Larisa Markeeva, Dylan Banarse, Skanda Koppula, Mateusz Malinowski, Yi Yang, Carl Doersch, et~al.
\newblock Perception test: A diagnostic benchmark for multimodal video models.
\newblock \emph{Advances in Neural Information Processing Systems}, 36, 2024.

\bibitem[Radford et~al.(2021)Radford, Kim, Hallacy, Ramesh, Goh, Agarwal, Sastry, Askell, Mishkin, Clark, et~al.]{radford2021learning}
Alec Radford, Jong~Wook Kim, Chris Hallacy, Aditya Ramesh, Gabriel Goh, Sandhini Agarwal, Girish Sastry, Amanda Askell, Pamela Mishkin, Jack Clark, et~al.
\newblock Learning transferable visual models from natural language supervision.
\newblock In \emph{International conference on machine learning}, pages 8748--8763. PMLR, 2021.

\bibitem[Rombach et~al.(2022)Rombach, Blattmann, Lorenz, Esser, and Ommer]{rombach2022high}
Robin Rombach, Andreas Blattmann, Dominik Lorenz, Patrick Esser, and Bj{\"o}rn Ommer.
\newblock High-resolution image synthesis with latent diffusion models.
\newblock In \emph{Proceedings of the IEEE/CVF conference on computer vision and pattern recognition}, pages 10684--10695, 2022.

\bibitem[Schuhmann et~al.(2022)Schuhmann, Beaumont, Vencu, Gordon, Wightman, Cherti, Coombes, Katta, Mullis, Wortsman, et~al.]{schuhmann2022laion}
Christoph Schuhmann, Romain Beaumont, Richard Vencu, Cade Gordon, Ross Wightman, Mehdi Cherti, Theo Coombes, Aarush Katta, Clayton Mullis, Mitchell Wortsman, et~al.
\newblock Laion-5b: An open large-scale dataset for training next generation image-text models.
\newblock \emph{Advances in Neural Information Processing Systems}, 35:\penalty0 25278--25294, 2022.

\bibitem[Sharma et~al.(2018)Sharma, Ding, Goodman, and Soricut]{sharma2018conceptual}
Piyush Sharma, Nan Ding, Sebastian Goodman, and Radu Soricut.
\newblock Conceptual captions: A cleaned, hypernymed, image alt-text dataset for automatic image captioning.
\newblock In \emph{Proceedings of the 56th Annual Meeting of the Association for Computational Linguistics (Volume 1: Long Papers)}, pages 2556--2565, 2018.

\bibitem[Singer et~al.(2022)Singer, Polyak, Hayes, Yin, An, Zhang, Hu, Yang, Ashual, Gafni, et~al.]{singer2022make}
Uriel Singer, Adam Polyak, Thomas Hayes, Xi Yin, Jie An, Songyang Zhang, Qiyuan Hu, Harry Yang, Oron Ashual, Oran Gafni, et~al.
\newblock Make-a-video: Text-to-video generation without text-video data.
\newblock \emph{arXiv preprint arXiv:2209.14792}, 2022.

\bibitem[Sun et~al.(2023{\natexlab{a}})Sun, Cui, Zhang, Zhang, Yu, Luo, Wang, Rao, Liu, Huang, et~al.]{sun2023emu2}
Quan Sun, Yufeng Cui, Xiaosong Zhang, Fan Zhang, Qiying Yu, Zhengxiong Luo, Yueze Wang, Yongming Rao, Jingjing Liu, Tiejun Huang, et~al.
\newblock Generative multimodal models are in-context learners.
\newblock \emph{arXiv preprint arXiv:2312.13286}, 2023{\natexlab{a}}.

\bibitem[Sun et~al.(2023{\natexlab{b}})Sun, Yu, Cui, Zhang, Zhang, Wang, Gao, Liu, Huang, and Wang]{sun2023emu}
Quan Sun, Qiying Yu, Yufeng Cui, Fan Zhang, Xiaosong Zhang, Yueze Wang, Hongcheng Gao, Jingjing Liu, Tiejun Huang, and Xinlong Wang.
\newblock Generative pretraining in multimodality.
\newblock \emph{arXiv preprint arXiv:2307.05222}, 2023{\natexlab{b}}.

\bibitem[Sun et~al.(2023{\natexlab{c}})Sun, Yu, Cui, Zhang, Zhang, Wang, Gao, Liu, Huang, and Wang]{sun2023generative}
Quan Sun, Qiying Yu, Yufeng Cui, Fan Zhang, Xiaosong Zhang, Yueze Wang, Hongcheng Gao, Jingjing Liu, Tiejun Huang, and Xinlong Wang.
\newblock Generative pretraining in multimodality.
\newblock \emph{arXiv preprint arXiv:2307.05222}, 2023{\natexlab{c}}.

\bibitem[Team(2024)]{team2024chameleon}
Chameleon Team.
\newblock Chameleon: Mixed-modal early-fusion foundation models.
\newblock \emph{arXiv preprint arXiv:2405.09818}, 2024.

\bibitem[Team et~al.(2023)Team, Anil, Borgeaud, Wu, Alayrac, Yu, Soricut, Schalkwyk, Dai, Hauth, et~al.]{team2023gemini}
Gemini Team, Rohan Anil, Sebastian Borgeaud, Yonghui Wu, Jean-Baptiste Alayrac, Jiahui Yu, Radu Soricut, Johan Schalkwyk, Andrew~M Dai, Anja Hauth, et~al.
\newblock Gemini: a family of highly capable multimodal models.
\newblock \emph{arXiv preprint arXiv:2312.11805}, 2023.

\bibitem[Team et~al.(2024)Team, Georgiev, Lei, Burnell, Bai, Gulati, Tanzer, Vincent, Pan, Wang, et~al.]{team2024gemini}
Gemini Team, Petko Georgiev, Ving~Ian Lei, Ryan Burnell, Libin Bai, Anmol Gulati, Garrett Tanzer, Damien Vincent, Zhufeng Pan, Shibo Wang, et~al.
\newblock Gemini 1.5: Unlocking multimodal understanding across millions of tokens of context.
\newblock \emph{arXiv preprint arXiv:2403.05530}, 2024.

\bibitem[Touvron et~al.(2023)Touvron, Lavril, Izacard, Martinet, Lachaux, Lacroix, Rozi{\`e}re, Goyal, Hambro, Azhar, et~al.]{touvron2023llama}
Hugo Touvron, Thibaut Lavril, Gautier Izacard, Xavier Martinet, Marie-Anne Lachaux, Timoth{\'e}e Lacroix, Baptiste Rozi{\`e}re, Naman Goyal, Eric Hambro, Faisal Azhar, et~al.
\newblock Llama: Open and efficient foundation language models.
\newblock \emph{arXiv preprint arXiv:2302.13971}, 2023.

\bibitem[Tschannen et~al.(2025)Tschannen, Eastwood, and Mentzer]{tschannen2025givt}
Michael Tschannen, Cian Eastwood, and Fabian Mentzer.
\newblock Givt: Generative infinite-vocabulary transformers.
\newblock In \emph{European Conference on Computer Vision}, pages 292--309. Springer, 2025.

\bibitem[Unterthiner et~al.(2018)Unterthiner, Van~Steenkiste, Kurach, Marinier, Michalski, and Gelly]{unterthiner2018towards}
Thomas Unterthiner, Sjoerd Van~Steenkiste, Karol Kurach, Raphael Marinier, Marcin Michalski, and Sylvain Gelly.
\newblock Towards accurate generative models of video: A new metric \& challenges.
\newblock \emph{arXiv preprint arXiv:1812.01717}, 2018.

\bibitem[Van Den~Oord et~al.(2017)Van Den~Oord, Vinyals, et~al.]{van2017neural}
Aaron Van Den~Oord, Oriol Vinyals, et~al.
\newblock Neural discrete representation learning.
\newblock \emph{Advances in neural information processing systems}, 30, 2017.

\bibitem[Wang et~al.(2024{\natexlab{a}})Wang, Suri, Ren, Chen, and Shrivastava]{wang2024larp}
Hanyu Wang, Saksham Suri, Yixuan Ren, Hao Chen, and Abhinav Shrivastava.
\newblock Larp: Tokenizing videos with a learned autoregressive generative prior.
\newblock \emph{arXiv preprint arXiv:2410.21264}, 2024{\natexlab{a}}.

\bibitem[Wang et~al.(2024{\natexlab{b}})Wang, Sun, Zhang, Tang, Liu, and Wang]{wang2024diffusion}
Wenxuan Wang, Quan Sun, Fan Zhang, Yepeng Tang, Jing Liu, and Xinlong Wang.
\newblock Diffusion feedback helps clip see better.
\newblock \emph{arXiv preprint arXiv:2407.20171}, 2024{\natexlab{b}}.

\bibitem[Wang et~al.(2024{\natexlab{c}})Wang, Yuan, Zhang, Chen, Wang, Zhang, Shen, Zhao, and Zhou]{wang2024videocomposer}
Xiang Wang, Hangjie Yuan, Shiwei Zhang, Dayou Chen, Jiuniu Wang, Yingya Zhang, Yujun Shen, Deli Zhao, and Jingren Zhou.
\newblock Videocomposer: Compositional video synthesis with motion controllability.
\newblock \emph{Advances in Neural Information Processing Systems}, 36, 2024{\natexlab{c}}.

\bibitem[Wang et~al.(2024{\natexlab{d}})Wang, Zhang, Luo, Sun, Cui, Wang, Zhang, Wang, Li, Yu, et~al.]{wang2024emu3}
Xinlong Wang, Xiaosong Zhang, Zhengxiong Luo, Quan Sun, Yufeng Cui, Jinsheng Wang, Fan Zhang, Yueze Wang, Zhen Li, Qiying Yu, et~al.
\newblock Emu3: Next-token prediction is all you need.
\newblock \emph{arXiv preprint arXiv:2409.18869}, 2024{\natexlab{d}}.

\bibitem[Wang et~al.(2023)Wang, He, Li, Li, Yu, Ma, Li, Chen, Chen, Wang, et~al.]{wang2023internvid}
Yi Wang, Yinan He, Yizhuo Li, Kunchang Li, Jiashuo Yu, Xin Ma, Xinhao Li, Guo Chen, Xinyuan Chen, Yaohui Wang, et~al.
\newblock Internvid: A large-scale video-text dataset for multimodal understanding and generation.
\newblock \emph{arXiv preprint arXiv:2307.06942}, 2023.

\bibitem[Wang et~al.(2024{\natexlab{e}})Wang, Xiong, Zhou, Lin, Zhao, Kang, Feng, and Liu]{wang2024loong}
Yuqing Wang, Tianwei Xiong, Daquan Zhou, Zhijie Lin, Yang Zhao, Bingyi Kang, Jiashi Feng, and Xihui Liu.
\newblock Loong: Generating minute-level long videos with autoregressive language models.
\newblock \emph{arXiv preprint arXiv:2410.02757}, 2024{\natexlab{e}}.

\bibitem[Wei et~al.(2023)Wei, Mangalam, Huang, Li, Fan, Xu, Wang, Xie, Yuille, and Feichtenhofer]{wei2023diffusion}
Chen Wei, Karttikeya Mangalam, Po-Yao Huang, Yanghao Li, Haoqi Fan, Hu Xu, Huiyu Wang, Cihang Xie, Alan Yuille, and Christoph Feichtenhofer.
\newblock Diffusion models as masked autoencoders.
\newblock In \emph{Proceedings of the IEEE/CVF International Conference on Computer Vision}, pages 16284--16294, 2023.

\bibitem[Wu et~al.(2021)Wu, Huang, Zhang, Li, Ji, Yang, Sapiro, and Duan]{wu2021godiva}
Chenfei Wu, Lun Huang, Qianxi Zhang, Binyang Li, Lei Ji, Fan Yang, Guillermo Sapiro, and Nan Duan.
\newblock Godiva: Generating open-domain videos from natural descriptions.
\newblock \emph{arXiv preprint arXiv:2104.14806}, 2021.

\bibitem[Wu et~al.(2024{\natexlab{a}})Wu, Chen, Wu, Ma, Liu, Pan, Liu, Xie, Yu, Ruan, et~al.]{wu2024janus}
Chengyue Wu, Xiaokang Chen, Zhiyu Wu, Yiyang Ma, Xingchao Liu, Zizheng Pan, Wen Liu, Zhenda Xie, Xingkai Yu, Chong Ruan, et~al.
\newblock Janus: Decoupling visual encoding for unified multimodal understanding and generation.
\newblock \emph{arXiv preprint arXiv:2410.13848}, 2024{\natexlab{a}}.

\bibitem[Wu et~al.(2023)Wu, Fei, Qu, Ji, and Chua]{wu2023nextgpt}
Shengqiong Wu, Hao Fei, Leigang Qu, Wei Ji, and Tat-Seng Chua.
\newblock Next-gpt: Any-to-any multimodal llm.
\newblock \emph{arXiv preprint arXiv:2309.05519}, 2023.

\bibitem[Wu et~al.(2024{\natexlab{b}})Wu, Zhang, Chen, Tang, Li, Fang, Zhu, Xie, Yin, Yi, et~al.]{wu2024vila}
Yecheng Wu, Zhuoyang Zhang, Junyu Chen, Haotian Tang, Dacheng Li, Yunhao Fang, Ligeng Zhu, Enze Xie, Hongxu Yin, Li Yi, et~al.
\newblock Vila-u: a unified foundation model integrating visual understanding and generation.
\newblock \emph{arXiv preprint arXiv:2409.04429}, 2024{\natexlab{b}}.

\bibitem[Xiang et~al.(2023)Xiang, Yang, Huang, and Wang]{xiang2023denoising}
Weilai Xiang, Hongyu Yang, Di Huang, and Yunhong Wang.
\newblock Denoising diffusion autoencoders are unified self-supervised learners.
\newblock In \emph{Proceedings of the IEEE/CVF International Conference on Computer Vision}, pages 15802--15812, 2023.

\bibitem[Xiao et~al.(2021)Xiao, Shang, Yao, and Chua]{xiao2021next}
Junbin Xiao, Xindi Shang, Angela Yao, and Tat-Seng Chua.
\newblock Next-qa: Next phase of question-answering to explaining temporal actions.
\newblock In \emph{Proceedings of the IEEE/CVF conference on computer vision and pattern recognition}, pages 9777--9786, 2021.

\bibitem[Xie et~al.(2024)Xie, Mao, Bai, Zhang, Wang, Lin, Gu, Chen, Yang, and Shou]{xie2024show}
Jinheng Xie, Weijia Mao, Zechen Bai, David~Junhao Zhang, Weihao Wang, Kevin~Qinghong Lin, Yuchao Gu, Zhijie Chen, Zhenheng Yang, and Mike~Zheng Shou.
\newblock Show-o: One single transformer to unify multimodal understanding and generation.
\newblock \emph{arXiv preprint arXiv:2408.12528}, 2024.

\bibitem[Xing et~al.(2025)Xing, Xia, Zhang, Chen, Yu, Liu, Liu, Wang, Shan, and Wong]{xing2025dynamicrafter}
Jinbo Xing, Menghan Xia, Yong Zhang, Haoxin Chen, Wangbo Yu, Hanyuan Liu, Gongye Liu, Xintao Wang, Ying Shan, and Tien-Tsin Wong.
\newblock Dynamicrafter: Animating open-domain images with video diffusion priors.
\newblock In \emph{European Conference on Computer Vision}, pages 399--417. Springer, 2025.

\bibitem[Xu et~al.(2016)Xu, Mei, Yao, and Rui]{xu2016msr}
Jun Xu, Tao Mei, Ting Yao, and Yong Rui.
\newblock Msr-vtt: A large video description dataset for bridging video and language.
\newblock In \emph{Proceedings of the IEEE conference on computer vision and pattern recognition}, pages 5288--5296, 2016.

\bibitem[Xu et~al.(2023)Xu, Liu, Vahdat, Byeon, Wang, and De~Mello]{xu2023open}
Jiarui Xu, Sifei Liu, Arash Vahdat, Wonmin Byeon, Xiaolong Wang, and Shalini De~Mello.
\newblock Open-vocabulary panoptic segmentation with text-to-image diffusion models.
\newblock In \emph{Proceedings of the IEEE/CVF Conference on Computer Vision and Pattern Recognition}, pages 2955--2966, 2023.

\bibitem[Xu et~al.(2024)Xu, Zhao, Zhou, Lin, Ng, and Feng]{xu2024pllava}
Lin Xu, Yilin Zhao, Daquan Zhou, Zhijie Lin, See~Kiong Ng, and Jiashi Feng.
\newblock Pllava: Parameter-free llava extension from images to videos for video dense captioning.
\newblock \emph{arXiv preprint arXiv:2404.16994}, 2024.

\bibitem[Yan et~al.(2021)Yan, Zhang, Abbeel, and Srinivas]{yan2021videogpt}
Wilson Yan, Yunzhi Zhang, Pieter Abbeel, and Aravind Srinivas.
\newblock Videogpt: Video generation using vq-vae and transformers.
\newblock \emph{arXiv preprint arXiv:2104.10157}, 2021.

\bibitem[Yang et~al.(2024{\natexlab{a}})Yang, Yin, Zhou, Rao, Zhai, Cao, and Zha]{yang2024mmar}
Jian Yang, Dacheng Yin, Yizhou Zhou, Fengyun Rao, Wei Zhai, Yang Cao, and Zheng-Jun Zha.
\newblock Mmar: Towards lossless multi-modal auto-regressive prababilistic modeling.
\newblock \emph{arXiv preprint arXiv:2410.10798}, 2024{\natexlab{a}}.

\bibitem[Yang et~al.(2024{\natexlab{b}})Yang, Ge, Li, Chen, Ge, Shan, and Chen]{yang2024seed}
Shuai Yang, Yuying Ge, Yang Li, Yukang Chen, Yixiao Ge, Ying Shan, and Yingcong Chen.
\newblock Seed-story: Multimodal long story generation with large language model.
\newblock \emph{arXiv preprint arXiv:2407.08683}, 2024{\natexlab{b}}.

\bibitem[Yi et~al.(2019)Yi, Gan, Li, Kohli, Wu, Torralba, and Tenenbaum]{yi2019clevrer}
Kexin Yi, Chuang Gan, Yunzhu Li, Pushmeet Kohli, Jiajun Wu, Antonio Torralba, and Joshua~B Tenenbaum.
\newblock Clevrer: Collision events for video representation and reasoning.
\newblock \emph{arXiv preprint arXiv:1910.01442}, 2019.

\bibitem[Yu et~al.(2023)Yu, Shi, Pasunuru, Muller, Golovneva, Wang, Babu, Tang, Karrer, Sheynin, et~al.]{yu2023scaling}
Lili Yu, Bowen Shi, Ramakanth Pasunuru, Benjamin Muller, Olga Golovneva, Tianlu Wang, Arun Babu, Binh Tang, Brian Karrer, Shelly Sheynin, et~al.
\newblock Scaling autoregressive multi-modal models: Pretraining and instruction tuning.
\newblock \emph{arXiv preprint arXiv:2309.02591}, 2023.

\bibitem[Yu et~al.(2024)Yu, Sun, Zhang, Cui, Zhang, Cao, Wang, and Liu]{yu2024capsfusion}
Qiying Yu, Quan Sun, Xiaosong Zhang, Yufeng Cui, Fan Zhang, Yue Cao, Xinlong Wang, and Jingjing Liu.
\newblock Capsfusion: Rethinking image-text data at scale.
\newblock In \emph{Proceedings of the IEEE/CVF Conference on Computer Vision and Pattern Recognition}, pages 14022--14032, 2024.

\bibitem[Yu et~al.(2019)Yu, Xu, Yu, Yu, Zhao, Zhuang, and Tao]{yu2019activitynet}
Zhou Yu, Dejing Xu, Jun Yu, Ting Yu, Zhou Zhao, Yueting Zhuang, and Dacheng Tao.
\newblock Activitynet-qa: A dataset for understanding complex web videos via question answering.
\newblock In \emph{Proceedings of the AAAI Conference on Artificial Intelligence}, pages 9127--9134, 2019.

\bibitem[Zhao et~al.(2024)Zhao, Song, Wang, Feng, Ding, Sun, Xiao, and Wang]{zhao2024monoformer}
Chuyang Zhao, Yuxing Song, Wenhao Wang, Haocheng Feng, Errui Ding, Yifan Sun, Xinyan Xiao, and Jingdong Wang.
\newblock Monoformer: One transformer for both diffusion and autoregression.
\newblock \emph{arXiv preprint arXiv:2409.16280}, 2024.

\bibitem[Zhao et~al.(2023)Zhao, Rao, Liu, Liu, Zhou, and Lu]{zhao2023unleashing}
Wenliang Zhao, Yongming Rao, Zuyan Liu, Benlin Liu, Jie Zhou, and Jiwen Lu.
\newblock Unleashing text-to-image diffusion models for visual perception.
\newblock In \emph{Proceedings of the IEEE/CVF International Conference on Computer Vision}, pages 5729--5739, 2023.

\bibitem[Zhou et~al.(2024)Zhou, Yu, Babu, Tirumala, Yasunaga, Shamis, Kahn, Ma, Zettlemoyer, and Levy]{zhou2024transfusion}
Chunting Zhou, Lili Yu, Arun Babu, Kushal Tirumala, Michihiro Yasunaga, Leonid Shamis, Jacob Kahn, Xuezhe Ma, Luke Zettlemoyer, and Omer Levy.
\newblock Transfusion: Predict the next token and diffuse images with one multi-modal model.
\newblock \emph{arXiv preprint arXiv:2408.11039}, 2024.

\bibitem[Zhou et~al.(2018)Zhou, Xu, and Corso]{zhou2018towards}
Luowei Zhou, Chenliang Xu, and Jason Corso.
\newblock Towards automatic learning of procedures from web instructional videos.
\newblock In \emph{Proceedings of the AAAI Conference on Artificial Intelligence}, 2018.

\bibitem[Zhu et~al.(2023)Zhu, Ding, Ge, Ge, Zhao, Zhao, Wang, and Shan]{zhu2023vl}
Jinguo Zhu, Xiaohan Ding, Yixiao Ge, Yuying Ge, Sijie Zhao, Hengshuang Zhao, Xiaohua Wang, and Ying Shan.
\newblock Vl-gpt: A generative pre-trained transformer for vision and language understanding and generation.
\newblock \emph{arXiv preprint arXiv:2312.09251}, 2023.

\end{thebibliography}
}
\clearpage
\appendix

\section{Implementation Details}
\subsection{Divot Tokenization.} 
{\flushleft \bf Model Architecture.} The Divot tokenizer is composed of a pre-trained ViT-H/14, a Spatial-Temporal Transformer and a Perceiver Resampler. Specifically, given a video clip with a duration of two seconds, we sample 5 frames at 2 fps, which are fed into the ViT to extract frame-level features. Subsequently, the extracted frame-level features are fed into the Spatial-Temporal Transformer, which consists of a 6-layer temporal transformer for temporal fusion, average pooling with a pool size of 5, and a 4-layer transformer for spatial and temporal fusion. To reduce the number of video tokens, these features after the Spatial-Temporal Transformer are further fed into the Perceiver Resampler, which contains 6-layer Perceiver Attention~\cite{alayrac2022flamingo}, for obtaining the final 64 video tokens for unified comprehension and generation with a LLM. We adopt the de-noising U-Net in DynamiCrafter~\cite{xing2025dynamicrafter} as the de-tokenizer, but reduce the input channel of the 3D convolution from 8 to 4 since we remove the original concatenation of a conditional image with noisy latents. To further enhance the reconstruction quality of the de-tokenizer, we add 6-layer Perceiver Attention~\cite{alayrac2022flamingo} after the Divot tokenizer to obtain 125 video tokens as the input of the U-Net, which are not used during the training of the LLM.

{\flushleft \bf Training Pipeline.} Since the original DynamiCrafter concatenates the conditional image
with per-frame initial noise and feeds them to the denoising
U-Net as a form of guidance, it cannot be directly applied to video representation learning due to its extra dependence on low-level image inputs. To address this, we first fine-tune the pre-trained DynamiCrafter by removing the concatenation of the conditional image. This modification makes the model utilize only the image and caption features, along with temporal embeddings, as the sole conditions for denoising the noisy video clips. Then we replace the image and caption features with spatiotemporal representations produced by Divot tokenizer as the conditions, and train the Divot tokenizer and the denoising U-Net in an end-to-end manner with $v$ prediction for denoising. After this stage, to further enhance the generation quality of our de-tokenizer, we freeze the Divot tokenizer and only fine-tune the denoising U-Net. During this fine-tuning process, we introduce a probability of 5\% to drop the conditions, enabling us to leverage classifier-free guidance during inference. Note that in previous stage for optimizing the Divot tokenizer, we do not drop conditions to ensure that the denoising process fully relies on the spatiotemporal representations to optimize representations.

{\flushleft \bf Training Data.} The Divot tokenizer is trained on pure videos of a subset of WebVid-10M~\cite{bain2021frozen} and Panda-70M~\cite{chen2024panda}, totaling 10M videos on 32 A100-40G GPUs. 
For WebVid-10M dataset, we employ LLaMA-3 to filter out videos with captions that do not contain dynamic content, resulting in a refined dataset of 4.8 million videos. For Panda-70M dataset, we download a total of 5.3 million videos, all of which are utilized for training purposes.

\subsection{Pre-training and Instruction Tuning.}
{\flushleft \bf Pre-training.}
Divot-LLM adopts next-word prediction and GMM modeling on video-text data for video comprehension and generation during pre-training. Specifically, the video features from the Divot tokenizer, the special tokens indicating the start and end of video features, along with the text tokens of the caption are fed into the pre-trained Mistral-7B~\cite{jiang2023mistral} for next token prediction trained with cross-entropy loss. Two fully-connected layers are trained to align the dimensions of the Divot features with those of the LLM. For GMM Modeling, text tokens of the caption and $N$ learnable queries are input into the LLM, where the output of the learnable queries are fed into two fully-connected layers to predict $2kd + k$ parameters per video token ($kd$ mean and $kd$ variance parameters for the mixture components, and $k$ mixture probabilities). We adopt $k=16$ in our experiment. We utilize bidirectional attention for $N$ learnable queries within the LLM and optimize the model using NLL loss. A total of 32 A100-40G GPUs are used for pre-training on 4.8 million video-caption pairs of WebVid-10M.

{\flushleft \bf Instruction Tuning.}
We perform multimodal instruction tuning on Divot-LLM to align it with human instructions
through supervised fine-tuning on public datasets as listed in Tab.~\ref{tab:data}. We fine-tune a LoRA module on the pre-trained Divot-LLM with the template as below,
\begin{equation}
\text{[INST] \quad\textless Instruction\textgreater\quad [/INST] \quad\textless Answer\textgreater}
\label{func:instruction}
\end{equation}

We further fine-tune the pretrained Divot-LLM on an animated series called ``Curious George'' to achieve video storytelling, which generates storyline and corresponding video clips in an interleaved  manner. Specifically, after downloading the videos of ``Curious George'' series, we adopt the video splitting algorithm in Panda-70M to cut a long video into several semantically coherent clips including splitting based on shot boundary detection, and stitching based on semantics similarity. 
Subsequently, we employ GPT-4V to generate captions for each video clip by uniformly sampling eight frames from each clip. Finally, we use GPT-4 to summarize the instructions and corresponding storylines based on the captions of three consecutive video clips.

After instruction tuning, to further enhance the quality of video generation, we adopt a de-tokenizer adaptation technique, which fine-tunes the de-tokenizer based on the features sampled from the predicted GMM distribution derived from the LLM output.

\begin{figure*}
	\centering
	\includegraphics[width=1.0\linewidth]{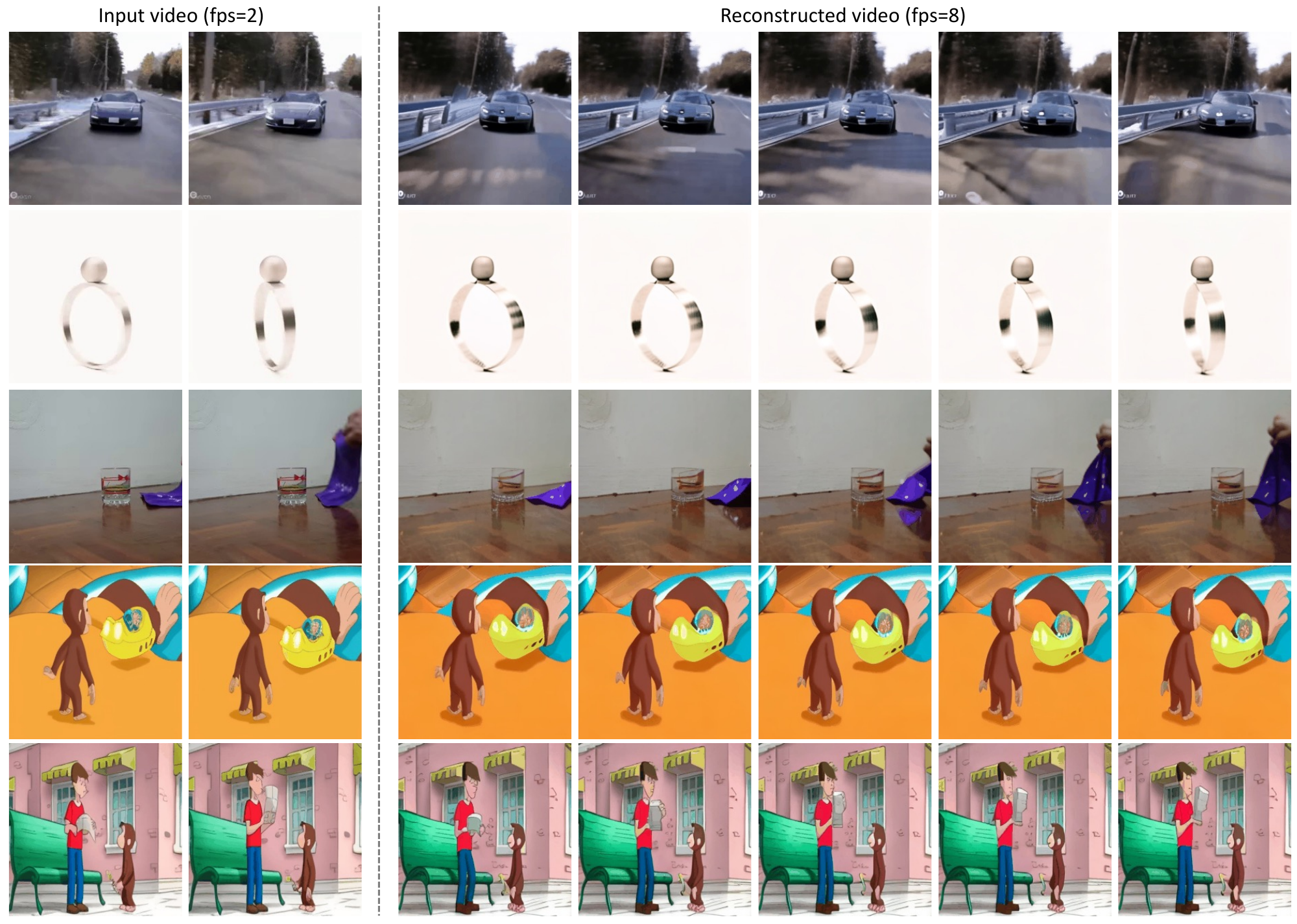}
	\caption{More qualitative examples of reconstructed videos, where the Divot tokenizer obtains spatiotemporal representations of sparsely sampled video frames and the de-tokenizer decodes these representations into semantically aligned and temporally coherent video clips.}
	\label{fig:reconstruction_supp}
\end{figure*}

\section{Qualitative Examples}

{\flushleft \bf Video Reconstruction.}
We provide additional qualitative examples of video reconstruction in Fig.~\ref{fig:reconstruction_supp}, where the spatiotemporal representations are obtained from the Divot tokenizer and subsequently fed into the denoising U-Net to denoise realistic video clips from noise. The decoded video clips, generated from the learned spatiotemporal representations, exhibit semantic alignment with the original videos and maintain temporal coherence. For the adaptation to the animated series "Curious George," we fine-tune only the de-tokenizer while keeping the Divot tokenizer frozen. The satisfactory reconstruction results demonstrate the generalizability of our Divot tokenizer in obtaining robust video representations.

\begin{figure*}
	\centering
	\includegraphics[width=1.0\linewidth]{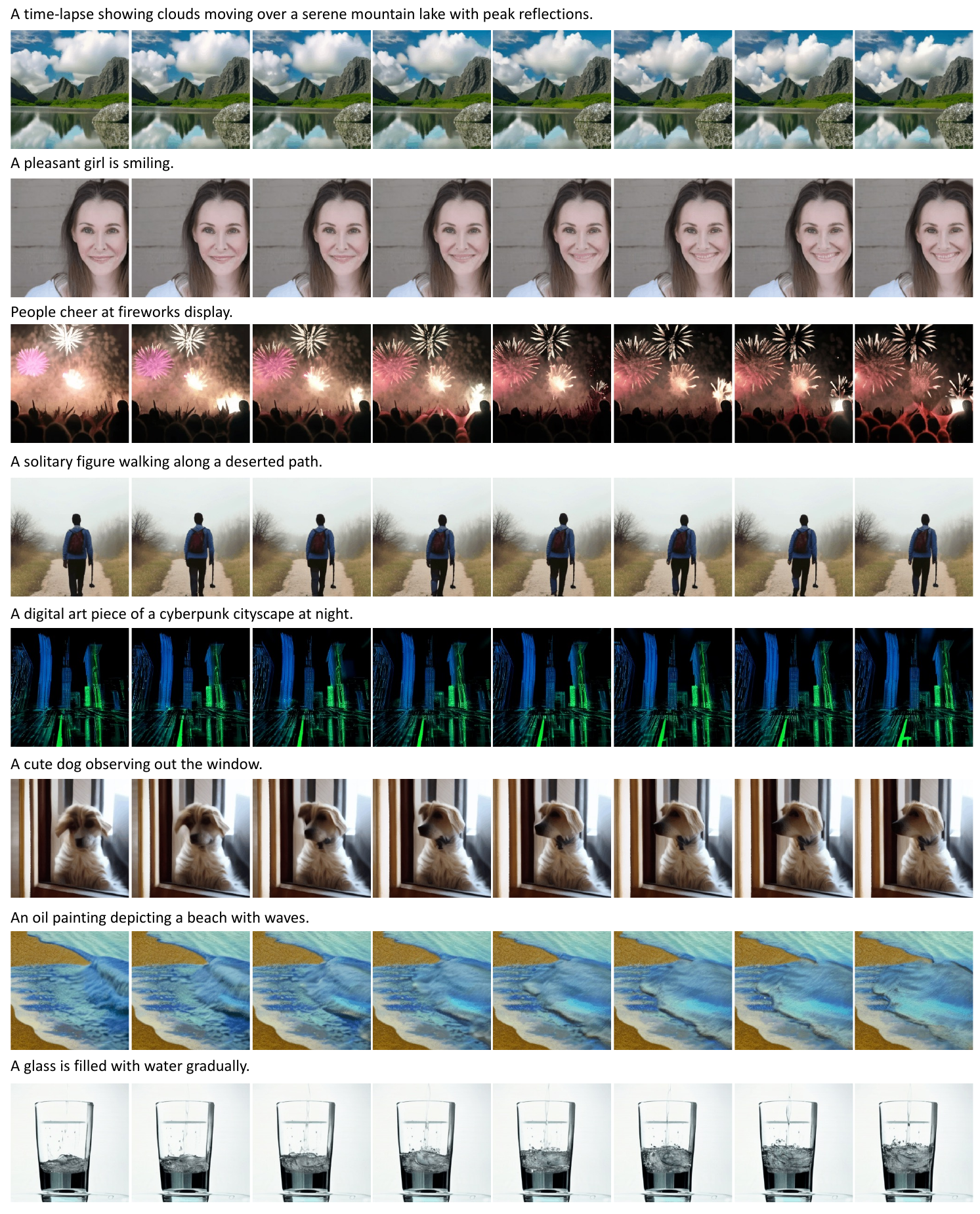}
	\caption{More qualitative examples of text-to-video generation by Divot-LLM, which effectively generates videos that are both semantically aligned with text prompts and temporally coherent across frames.}
	\label{fig:video_gen_supp}
	\vspace{-2mm}	
\end{figure*}

{\flushleft \bf Video Generation.}
We present more qualitative examples of text-to-video generation in Fig.~\ref{fig:video_gen_supp}. Through modeling the distributions of Divot features with a GMM and training the LLM to predict GMM parameters, our Divot-LLM can generate videos that are both semantically aligned with text prompts and temporally coherent across frames. This is achieved through the dual-function de-tokenizer, utilizing only 4.8 million video-caption pairs for training.

\begin{figure*}
	\centering
	\includegraphics[width=1.0\linewidth]{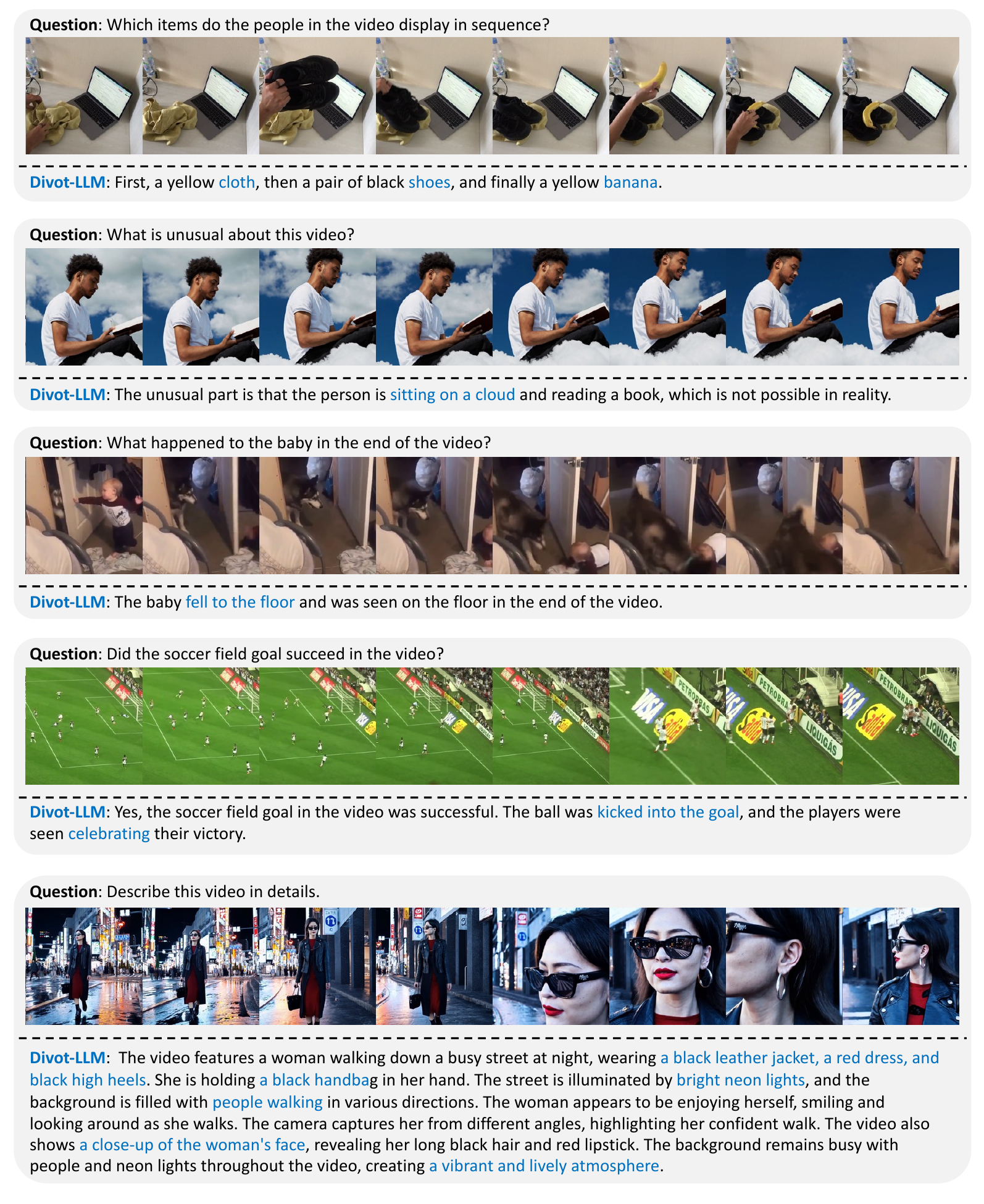}
	\caption{Qualitative examples of video comprehension by Divot-LLM.}
	\label{fig:video_comp_supp}
	\vspace{-2mm}	
\end{figure*}

{\flushleft \bf Video Comprehension.}
As illustrated in Fig.~\ref{fig:video_comp_supp}, we provide qualitative examples to demonstrate the video comprehension capability of Divot-LLM. It can effectively understand sequences of events depicted in a video, reason using common sense, track and summarize the outcomes of specific actions or events, and deliver comprehensive and detailed descriptions of the videos. By utilizing diffusion procedure for video representation learning, our Divot tokenizer effectively captures robust spatiotemporal representations, enhancing the comprehension capabilities of Divot-LLM.

\end{document}